# Probability Bracket Notation: Markov Sequence Projector of Visible and Hidden Markov Models in Dynamic Bayesian Networks

Xing M. Wang[1]

## Abstract

With the symbolic framework of Probability Bracket Notation (PBN), the Markov Sequence Projector (MSP) is introduced to expand the evolution formula of Homogeneous Markov Chains (HMCs). The well-known weather example, a Visible Markov Model (VMM), illustrates that the full joint probability of a VMM corresponds to a specifically projected Markov state sequence in the expanded evolution formula. In a Hidden Markov Model (HMM), the probability basis (P-basis) of the hidden Markov state sequence and the P-basis of the observation sequence exist in the sequential event space. The full joint probability of an HMM is the product of the (unknown) projected hidden sequence of Markov states and their transformations into the observation P-bases. The Viterbi algorithm is applied to the famous Weather-Stone HMM example to determine the most likely weather-state sequence given the observed stone-state sequence. Our results are verified using the Elvira software package. Using the PBN, we unify the evolution formulas for Markov models like VMMs, HMMs, and factorial HMMs (with discrete time). We briefly investigated the extended HMM, addressing the feedback issue, and the continuous-time VMM and HMM (with discrete or continuous states). All these models are subclasses of Dynamic Bayesian Networks (DBNs) essential for Machine Learning (ML) and Artificial Intelligence (AI).

[1] Sherman Visual Lab, Sunnyvale, CA 94085, USA, xmwang@shermanlab.com

## 1. Introduction: the PBN and the Discrete Markov Chains

Inspired by the remarkable success of Dirac's notation in quantum physics, we proposed the Probability Bracket Notation (PBN) [1]. We applied PBN to analyze Homogeneous Markov chains (HMC) [1, Sec 3] [2, Chapter 11]. This article focuses on the HMCs that operate in discrete time and have a finite number of discrete states (except in Section 6).

### 1.1. Time-independent Discrete Random Variables in the PBN

First, let's examine a time-independent discrete random variable. We assume its sample space Ω contains *N* states (or outcomes) denoted by $\Omega = \{s_1, ..., s_i, \ldots, s_N\}$. Utilizing Eq. (19) of [1], we derive the following time-independent probability basis (P-basis):

Orthogonality: $$P(i\,|\,j) \equiv P(s_i\,|\,s_j) = \delta_{ij} \tag{1.1.1a}$$

Completeness: $$\sum_{i=1}^{N} |\,i)P(i\,| \equiv \sum_{i=1}^{N} |\,s_i)P(s_i\,| = I_S \quad \text{(P-identity)} \tag{1.1.1b}$$

As described in Ref [1], there is a one-to-one map between the above P-basis and the V-basis in Hilbert space using the Dirac notation:

Orthogonality: $$\langle i\,|\,j\rangle \equiv \langle s_i\,|\,s_j\rangle = \delta_{ij} \tag{1.1.2a}$$

Completeness: $$\sum_{i=1}^{N} |\,i\rangle\langle i\,| \equiv \sum_{i=1}^{N} |\,s_i\rangle\langle s_i\,| = I_S \quad \text{(V-identity)} \tag{1.1.2b}$$

Inserting the P-identity (or P-unit) operator $I_S$, one can show that the sum of probabilities for all states equals one, as required by the normalization [1, Eq. (22)]:

$$1 = P(\Omega\,|\,\Omega) = P(\Omega\,|\,I_S\,|\,\Omega) = P(\Omega\,|\sum_{i=1}^{N}|\,i)P(i\,|\,\Omega) = \sum_{i=1}^{N} P(s_i\,|\,\Omega) \tag{1.1.3}$$

$$\therefore \sum_{i=1}^{N} m_i = 1, \quad \text{where: } m_i \equiv P(i\,|\,\Omega) \equiv P(i) \tag{1.1.4}$$

In deriving Eq. (1.1.3), we have used the property of the conditional probability:

$$P(\Omega \mid i) = 1, \quad \because \Omega \supset s_i \neq \varnothing \tag{1.1.5}$$

**The Weather Example**: The simplest and most commonly utilized HMC example is the 3-state Markov model of weather ([2], [5] or [6]). We assume that there are three types of weather:

$$S = \{\text{sunny, rainy, foggy}\} = \{S, R, F\} = \{s_1, s_2, s_3\} \tag{1.1.6}$$

The weather is observed once a day, and it falls into one of the three states mentioned above. By monitoring the weather over several days, we establish the following time-independent probability distribution (PD):

$$\begin{aligned} &\pi_1 = P(sunny) = P(s_1 \mid \Omega) = P(1 \mid \Omega) = P(s_1) = P(1) = 0.5, \\ &\pi_2 = P(rainy) = P(2) = 0.3, \quad \pi_3 = P(foggy) = P(3) = 0.2 \end{aligned} \tag{1.1.7}$$

In this context, "time independence" refers to the PD remaining constant regardless of the specific day you observe it.

The orthogonality relation in Eq. (1.1.1) now reads:

$$\begin{aligned} &P(sunny \mid sunny) = P(1 \mid 1) = 1, \quad P(rainy \mid rainy) = 1, \quad P(foggy \mid foggy) = 1 \\ &P(sunny \mid rainy) = P(1 \mid 2) = 0, \quad P(sunny \mid foggy) = P(1 \mid 3) = 0, \ldots \end{aligned} \tag{1.1.8}$$

The completeness (or the P-Identity Operator) in Eq. (1.1.2) reads:

$$\sum_{i=1}^{3} |i)P(i| = sunny)P(sunny| + |rainy)P(rainy| + \quad |foggy)P(foggy| = I_S \tag{1.1.9}$$

The normalization Eq. (1.1.4) reads:

$$\sum_{i=1}^{3} \pi_i = \sum_{i=1}^{3} P\,(i) = P(sunny) + P(rainy) + P(foggy) = 1 \tag{1.1.10}$$

## 1.2. The Markov Evolution Formula of Markov Chains in the PBN

The sample space of HMCs has the following $N$-dimensional $P$-basis [1, §3.1]:

$$P(i,t \mid j,t) \equiv P(s_i,t \mid s_j,t) = \delta_{ij}, \quad \sum_{i=1}^{N} |\, i,t)P(i,t \,| \equiv \sum_{i=1}^{N} |\, s_i,t)P(s_i,t \,| = I_S(t) \tag{1.2.1a}$$

$$s_j \in S = \{s_1, \ldots, s_i, \ldots, s_N\} \tag{1.2.1b}$$

They also possess the corresponding V-basis in Hilbert space with the Dirac notation:

Orthogonality: $\langle i,t\,|\,j,t\rangle \equiv \langle s_i,t\,|\,s_j,t\rangle = \delta_{ij}$ (1.2.2a)

Completeness: $\sum_{i=1}^{N} |\,i,t\rangle\langle i,t\,| \equiv \sum_{i=1}^{N} |\,s_i,t\rangle\langle s_i,t\,| = I_S(t)$ (1.2.2b)

Note that all *P*-bras and *P*-kets are taken at the same time *t*.

The transition matrix is denoted by $\boldsymbol{A} = \{\, a_{i,\,j} \,\}$, defined as the transition probability from state *i* at time *t* to state *j* at time *t* + 1 (where *t* is an integer, representing steps). Its elements are non-negative real numbers and are time-independent for an HMC [4, 8]:

$$a_{i,\,j} \equiv P(q_{t+1} = j \,|\, q_t = i) \equiv P(j,t+1\,|\,i,t) = P(j,1\,|\,i,0), \quad t = \{1,2,...T\} \tag{1.2.3}$$

In the final step, we utilized the homogeneous property. They also adhere to the following stochastic constraints:

$$a_{i,\,j} \ge 0, \quad \sum_{j=1}^{N} a_{i,\,j} = 1, \quad 1 \le i,\, j \le N \tag{1.2.4}$$

To avoid possible confusion of indices, we denote:

$$a_{q_t,\,q_{t+1}} \equiv P(q_{t+1}\,|\,q_t), \quad q_t \equiv q(t):\; t \to s \in S \tag{1.2.5}$$

Therefore:

$$a_{i,\,j} \equiv P(q_{t+1} = j \,|\, q_t = i) \equiv a_{q_t = i,\; q_{t+1} = j} \tag{1.2.6}$$

Note that the indices *i* and *j* of $a_{i,\,j}$ are used to label state, while the indices *i* and *j* in *q*(*i*) and *q*(*j*) are used to label time. They have quite different meanings.

In the PBN [1, Eq. (55)], the system probability vector has the following evolution expression in Hilbert space:

$$|\,\Omega^{(t)}\rangle = \tilde{A}\,|\,\Omega^{(t-1)}\rangle = \tilde{A}^{t-1}\,|\,\Omega^{(1)}\rangle, \quad 1 \le t \le T \qquad (\tilde{A} \text{ is the transport of } A) \tag{1.2.7}$$

Mapped to the probability space, the evolution of the system *P*-ket of an HMC reads:

$$|\,\Omega_t) = \tilde{A}\,|\,\Omega_{t-1}) = \tilde{A}^{t-1}\,|\,\Omega_1), \quad 1 \le t \le T \qquad (\tilde{A} \text{ is the transport of } A) \tag{1.2.8}$$

Eq. (1.2.8) defines the *Markov evolution formula* (MEF). It is an expression in the Schrodinger picture because the system *P*-ket is explicitly time-dependent [1, §3.3]. The initial state can be expanded using the identity operator $I_S(t=1)$ in (1.2.1a):

$$|\,\Omega_1) = I_S(t=1)\,|\,\Omega_1) = \sum_{i=1}^{N} |\,i,1)P(i,1|\Omega_1) = \quad \sum_{i=1}^{N} \pi_i \,|\, i, t=1) \tag{1.2.9}$$

Here, the initial probability distribution is represented by vector $\boldsymbol{\pi}$ as in Eq. (1.1.7);

$$\pi_i = P(q_1 = s_i) = P(s_i, t=1\,|\,\Omega_1) \equiv P(i,1\,|\,\Omega_1), \quad \sum_{i=1}^{N} \pi_i = 1 \tag{1.2.10}$$

The probability distribution $P^{(t)}{}_i$ (at time $t$) is obtained by using the $I_S(t)$ in Eq. (1.2.1a):

$$|\,\Omega_t) = I_S(t)\,|\,\Omega_t) = \sum_{i}^{N} |\,i,t)P(i,t\,|\,\Omega_t) = \sum_{i}^{N} p^{(t)}{}_i \,|\, i,t), \quad \sum_{i}^{N} P(i,t\,|\,\Omega_t) \equiv \sum_{i}^{N} p^{(t)}{}_i = 1 \tag{1.2.11}$$

One can easily get the matrix representation of (1.2.7), as discussed in [1, §3.2]:

$$\begin{bmatrix} p^{(t+1)}{}_1 \\ \vdots \\ p^{(t+1)}{}_N \end{bmatrix} = \begin{pmatrix} a_{1,1} & \cdots & a_{N,1} \\ \vdots & \ddots & \vdots \\ a_{1,N} & \cdots & a_{N,N} \end{pmatrix} \begin{bmatrix} p^{(t)}{}_1 \\ \vdots \\ p^{(t)}{}_N \end{bmatrix}, \qquad \begin{bmatrix} p^{(1)}{}_1 \\ \vdots \\ p^{(1)}{}_N \end{bmatrix} = \begin{bmatrix} \pi_1 \\ \vdots \\ \pi_N \end{bmatrix} \tag{1.2.12}$$

Or, utilizing the notations of probability vectors in Dirac's notation:

$$|\,p^{(t+1)}\rangle = \tilde{A}\cdot|\,p^{(t)}\rangle, \quad |\,p^{(1)}\rangle = |\,\pi\rangle, \quad \text{where } \langle i\,|\,p^{(t)}\rangle = p^{(t)}{}_i, \quad \langle i\,|\,\pi\rangle = \pi_i \tag{1.2.13}$$

Here is the static V-basis of the Markov states in Hilbert space:

$$\langle s_i\,|\,s_j\rangle \equiv \langle i\,|\,j\rangle = \delta_{i,j} \qquad \sum_{i=1}^{N} |\,s_i\rangle\langle s_i\,| \equiv \sum_{i=1}^{N} |\,i\rangle\langle i\,| = I_S \tag{1.2.14}$$

$$\langle s_i\,|\,\tilde{A}\,|\,s_j\rangle \equiv \langle i\,|\,\tilde{A}\,|\,j\rangle = \tilde{a}_{i,j} = a_{j,i} \tag{1.2.15}$$

Combining the above equations, we obtain the following expression in Dirac's notation:

$$\langle q_T\,|\,\tilde{A}^{T-1}\,|\,q_1\rangle = \sum_{q_2,\cdots q_{T-1}} a_{q_{T-1},\,q_T} \cdot a_{q_{T-2},\,q_{T-1}} \cdots a_{q_1,\,q_2} \tag{1.2.16}$$

In the probability space, the transition operator $A$ is defined by:

$$A \equiv \sum_{i,k} |\,s_i)\; a_{i,k} P(s_k\,| \equiv \sum_{i,k} |\,i)\; a_{i,k} P(k\,| \tag{1.2.17a}$$

$$\tilde{A} \equiv \sum_{i,k} |\,s_i)\; a_{k,i} P(s_k\,| \equiv \sum_{i,k} |\,i)\; a_{k,i} P(k\,| \tag{1.2.17b}$$

Applying Eq. (1.1.1), we have:

$$P(s_i \mid \tilde{A} \mid s_j) \equiv P(i \mid \tilde{A} \mid j) = a_{j,i} \quad \text{or} \quad P(q_{t+1} \mid \tilde{A} \mid q_t) = a_{q_t, q_{t+1}} \tag{1.2.18}$$

Eq. (1.2.8) and (1.2.10) lead to the *expanded Markov Evolution Formula*:

$$P(q_T \mid \Omega_T) = P(q_T \mid \tilde{A}^{T-1} \mid \Omega_1) = \sum_{q_1, \cdots q_{T-1}} a_{q_{T-1}, q_T} \cdot a_{q_{T-2}, q_{T-1}} \cdots a_{q_1, q_2} \cdot \pi_{q_1} \tag{1.2.19}$$

$$\mid \Omega_T) = \tilde{A} \mid \Omega_{T-1}) = \tilde{A}^{T-1} \mid \Omega_1) \tag{1.2.20}$$

## 1.3. The Weather Example and the Transition Matrix

Let's now assume that, after observing the weather for several days, we arrive at the following transitions [6]:

| Today's Weather | Tomorrow's Weather | | |
|---|---|---|---|
| | Sunny | Rainy | Foggy |
| Sunny | 0.8 | 0.05 | 0.15 |
| Rainy | 0.2 | 0.6 | 0.2 |
| Foggy | 0.2 | 0.3 | 0.5 |

**Table 1.2.1**: The Weather Transitions

From Table (1.2.1), we derive the weather transition matrix [6]:

$$\boldsymbol{A} = \{a_{i,j}\} = \{P(q_{t+1} = j \mid q_t = i)\} = \begin{pmatrix} 0.8 & 0.05 & 0.15 \\ 0.2 & 0.6 & 0.2 \\ 0.2 & 0.3 & 0.5 \end{pmatrix} \tag{1.3.1}$$

Let us examine what we can and cannot do with our expanded MEF, Eq. (1.2.19). We denote the state observed on day t by $q_t$ to ensure consistency.

**Problem 1.3.1**: Given that today the weather is sunny ($q_1 = s_1$), what is the probability that the day after tomorrow is rainy ($q_3 = s_2$)?

From the condition given, we know that $T = 3$ and $\pi_i = \delta_{i,1}$. Applying them to Eq. (1.2.19), we have:

$$\begin{aligned} P(q_2 \mid \Omega_3) &= P(2 \mid \tilde{A}^2 \mid \Omega_1) = \sum_{q_1,\ q_2} a_{q_2, 2} \cdot a_{q_1, q_2} \cdot \pi_{q_1} \\ &= \sum_k a_{k,2} \cdot a_{1,k} = \{A^2\}_{1,2} \underset{(1.2.20)}{=} 0.115 \end{aligned} \tag{1.3.2}$$

**Problem 1.3.2**: Given that today, the weather is sunny ($q_1 = s_1$), what is the probability that tomorrow will be foggy ($q_2 = s_3$) and the day after tomorrow will be rainy ($q_3 = s_2$)?

We cannot answer the question using Eq. (1.2.19) because it only produces the probability distribution at the final time $t = 3$, given the initial probability distribution at the initial time $t = 1$. But, we can find the solution by specifying the intermediate states:

$$\begin{aligned} &P(q_1 = 1, q_2 = 3, q_3 = 2 \mid \Omega_1) = a_{3,2} \cdot a_{1,3} \cdot \pi_1 \\ &= a_{1,3} \cdot a_{3,2} \cdot \delta_{1,1} = 0.15 \times 0.3 = 0.045 \end{aligned} \tag{1.3.3}$$

We refer to this type of Markov model as the *Visible Markov Model* (VMM), which addresses the full joint probability (FJP) of a specified path in the expanded MEF of HMC (1.2.19):

$$P(Q_T \mid \Omega_1) \equiv P(q_1, q_2, \ldots q_T \mid \Omega_1) = a_{q_{T-1}, q_T} \cdot a_{q_{T-2}, q_{T-1}} \cdots a_{q_1, q_2} \cdot \pi_{q_1} \tag{1.3.4}$$

Here $Q_T$ represents the special Markov state sequence (MSS):

$$Q_T \equiv Q_{[1:T]} \equiv \{q_1, q_2, \ldots, q_T\}, \quad T \geq 1 \tag{1.3.5}$$

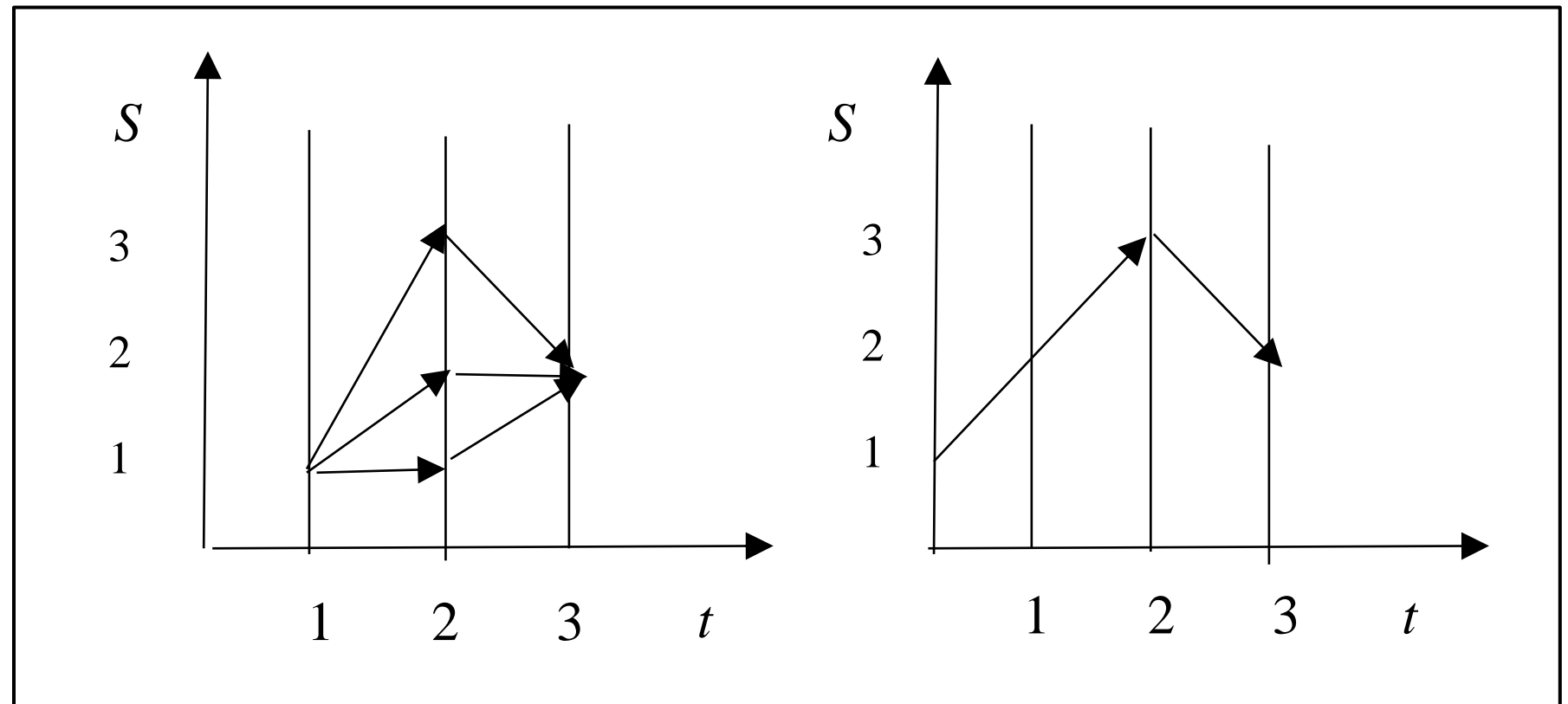

**Fig. 1.3.1**. HMC vs VMM; Left: Eq. (1.3.2); Right: Eq. (1.3.3)

Fig. (1.3.1) shows the difference of HMC and VMM: HMC has all possible intermediate state sequences (paths) while VMM has only a specified sequence (path)[2].

### 1.4. The P-projectors and the Markov Sequence Projector

To prepare for studying VMM, we define several operators related to the time evolution for a discrete Markov state sequence (MSS) using the PBN. The first operator is the single-state *probability projector* (P-projector):

$$U(q_t) \equiv |\, q_t) P(q_t \,|, \quad U(q_t = j) \equiv U(j,t) = |\, j,t) P(j,t \,|, \quad U(q_t) U(q_t) = U(q_t) \tag{1.4.1}$$

[2] In our discussion, we refer to an HMC if its transition probability is described by Eq. (1.2.19) or (1.4.7) and refer to a VMM if it is described by Eq. (1.3.4).

The P-projectors are already utilized in P-identities, as shown in Eq. (1.2.1a). The product of two consecutive P-projectors represents a transition operator:

$$U(q_t)U(q_{t-1}) = |\, q_t)P(q_t \,|\, q_{t-1})P(q_{t-1}\,| = |\, q_t)a_{q_{t-1},q_t}P(q_{t-1}\,|$$

For a given state sequence in Eq. (1.3.5), the *Markov Sequence Projector* (MSP) is defined by the product of successive P-projectors of the MSS, Eq. (1.3.5):

$$\begin{aligned} U(q_{[T:1]}) &\equiv \prod_{\tau=1}^{T-1} U(q_{T-\tau}) = U(q_{T-1})\cdots U(q_1) \\ &= |\, q_{T-1})P(q_{T-1}\,|\, q_{T-2})...P(q_2\,|\, q_1)P(q_1\,| = |\, q_{T-1})\, a_{q_{T-2},\, q_{T-1}} \cdots a_{q_1,\, q_2} P(q_1\,| \end{aligned} \tag{1.4.2}$$

The MSP projects a specific MSS, as shown in Fig. (1.3.1). It is not a unitary operator but exhibits the necessary behavior for a time evolution operator.

$$U(q_{[T:1]}) = U(q_{T-1})U(q_{[T-2:1]}) = U(q_{[T:t]})U(q_{[t:1]}), \quad 1 < t \le T \tag{1.4.3}$$

Applying Eq. (1.2.5) and Eq. (1.2.10), for an MSS $Q_T$, Eq. (1.4.2) leads to:

$$P(q_T\,| U(q_{[T:1]}) \,|\, \Omega_1) = a_{q_{T-1},\, q_T} a_{q_{T-2},\, q_{T-1}} \cdots a_{q_1,\, q_2} \cdot \pi_{q_1} \tag{1.4.4}$$

Its right side is identical to the FJP for VMM, Eq. (1.3.4), that is

$$P(Q_T\,|\,\Omega_1) \equiv P(q_1, q_2, \ldots q_T\,|\,\Omega_1) = P(q_T\,|U(q_{[T:1]})\,|\,\Omega_1) \equiv P(q_T\,|\Omega_T) \tag{1.4.5}$$

$$|\,\Omega_T) = U(q_{T-1})\,|\,\Omega_{T-2}) = U(q_{[T:1]})\,|\,\Omega_1) \tag{1.4.6}$$

With Eq. (1.4.5), the expanded MEF of HMC, Eq. (1.2.19), now can be represented as the sum of all possible paths in the VMM:

$$\begin{aligned} P(q_T\,|\,\Omega_T) &= P(q_T\,|\,\tilde{A}^{T-1}\,|\,\Omega_1) = \sum_{q_1,\cdots q_{T-1}} P(q_T\,|U(q_{[T:1]})\,|\,\Omega_1) \\ &= \sum_{q_1,\cdots q_{T-1}} P(q_T\,|U(q_{T-1})\cdots U(q_1)\,|\,\Omega_1) \equiv \sum_{q_1,\cdots q_{T-1}} P(q_1, q_2, ..., q_T\,|\,\Omega_1) = \sum_{q_1,\cdots q_{T-1}} P(Q_T\,|\,\Omega_1) \end{aligned} \tag{1.4.7}$$

Eq. (1.2.8) can now be expressed as:

$$|\,\Omega_T) = \tilde{A}^{T-1}\,|\,\Omega_1) = \sum_{q_1,\cdots q_{T-1}} U(q_{[T:1]})\,|\,\Omega_1) \tag{1.4.8}$$

$$\tilde{A}^{T-1} = \sum_{q_1,\cdots q_{T-1}} U(q_{[T:1]}) = \sum_{q_1,\cdots q_{T-1}} |\, q_{T-1})\, a_{q_{T-2},\, q_{T-1}} \cdots a_{q_1,\, q_2} P(q_1\,| \tag{1.4.9}$$

Eq. (1.4.8) represents the time evolution operator for the system *P*-ket of an HMC.

## 2. The PBN and Visible Markov Models

Let us examine questions similar to Problems 1.3.1 and 1.3.2 using Eq. (1.3-4). Because all these questions [6] are related to a Markov state sequence, our answers can serve as an introduction to the Visible Markov Model (VMM).

### 2.1. The VMM and the Weather Example

**Problems 2.1.1:** Given that today the weather is sunny ($q_1 = s_1$), what is the probability that tomorrow will be sunny ($q_2 = s_1$) and the day after will be rainy ($q_3 = s_2$)?

Because today is sunny, we have $P(q_1 = s_1) = \pi_1 = 1$, so Eq. (1.2.10) now reads:

$$[\pi_1, \pi_2, \pi_3] = [1, 0, 0], \quad \text{or: } \pi_i = \delta_{1,i} \tag{2.1.1}$$

The probability that tomorrow is also sunny can be calculated from (1.2.12) as:

$$p^{(2)}{}_1 = P(q_2 = 1 \mid q_1 = 1) = \sum_{i=1}^{3} a_{i,1} \cdot \pi_i = \sum_{i=1}^{3} a_{i,1} \cdot \delta_{i,1} = a_{1,1} = 0.8$$

Given that tomorrow is sunny, the probability that the day after is rainy is given by:

$$p^{(3)}{}_2 = P(q_3 = 2 \mid q_2 = 1) = \sum_{i=1}^{3} a_{i,2} \cdot \delta_{i,1} = a_{1,2} = 0.05$$

Combining all together, we can write the joint probability of the state sequence:

$$\begin{aligned} &P(q_3 = 2, q_2 = 1, q_1 = 1 \mid \Omega_1) = P(q_3 = 2 \mid q_2 = 1) \cdot P(q_2 = 1 \mid q_1 = 1) \cdot P(q_1 \mid \Omega_1) \\ &= a_{1,2} \cdot a_{1,1} \cdot \pi_1 = 0.05 \cdot 0.8 \cdot 1 = 0.04 \end{aligned} \tag{2.1.2}$$

Eq. (2.1.2) can be expressed symbolically as:

$$P(q_3, q_2, q_1 \mid \Omega_1) = P(q_3 \mid q_2) \cdot P(q_2 \mid q_1) \cdot P(q_1 \mid \Omega_1) = a_{q_2, q_3} \cdot a_{q_1, q_2} \cdot \pi_{q_1} \tag{2.1.3}$$

Moreover, using the MSP in Eq. (1.4.2), we can rewrite Eq. (2.1.3) for the Markov state sequence $Q_3 = Q_{[1:3]} = \{q_1, q_2, q_3\}$ as follows:

$$P(q_1, q_2, q_3 \mid \Omega_1) = P(q_3 \mid q_2) P(q_2 \mid q_1) P(q_1 \mid \Omega_1) = P(q_3 \mid U(q_{[3:1]}) \mid \Omega_1) \tag{2.1.4}$$

We have now unified all following expressions for the same FJP:

$$P(Q_{[1:3]} \mid \Omega_1) = P(q_1, q_2, q_3 \mid \Omega_1) = P(q_3 \mid U(q_{[3:1]}) \mid \Omega_1) = a_{q_2, q_3} \cdot a_{q_1, q_2} \cdot \pi_{q_1} \tag{2.1.5}$$

**Problems 2.1.2:** Assume that the weather yesterday was rainy ($q_1 = s_2$), and today it is foggy ($q_2 = s_3$); then what is the probability that it will be sunny ($q_3 = s_1$) tomorrow? Using the Markov property and the condition $P(q_2 = 3|\Omega_2) = P(3|\Omega_2) = 1$, we find:

$$P(q_3 = 1, q_2 = 3) = P(1|3)\cdot P(3|\Omega_2) = a_{3,1} = 0.2 \tag{2.1.6}$$

**Problems 2.1.3:** Given that the weather today is sunny ($q_1 = s_1$), what is the probability that the day after is rainy ($q_3 = s_2$)? This is the same question as Problem 1.3.1**.** Because $q_2$ can be any state, we get the result by summing over $q_2 = i$ in Eq. (2.1.3):

$$\begin{aligned}&\sum_{i=1}^{3} P(q_3, i, q_1) = \sum_{i=1}^{3} P(q_3 = 2|q_2 = i)\cdot P(q_2 = i|q_1 = 1)\cdot P(q_1 = 1|\Omega_1)\\ &= \sum_{i=1}^{3} a_{i2}\cdot a_{1i}\cdot \pi_1 == \{A^2\}_{1,2} \underset{(1.3.2)}{=} 0.115\end{aligned} \tag{2.1.7}$$

The answer is identical to Eq. (1.3.2), obtained using Eq. (1.2.8), the Markov evolution formula for system *P*-ket of a HMC. We can extend Eq. (2.1.5) to a general Markov state sequence (MSS) $Q_T$:

$$P(Q_T|\Omega_1) \equiv P(q_1, q_2, ..., q_T|\Omega_1) = P(q_T|U(q_{[T:1]})\Omega_1) = P(q_T|q_{T-1})\cdots P(q_1|\Omega_1) \tag{2.1.8}$$

Using the transition matrix $\boldsymbol{A}$ and the initial condition $\boldsymbol{\pi}$, we obtain the FJP for MSS $Q_T$:

$$\begin{aligned}&P(Q_T|\Omega_1) \equiv P(q_1, q_2, \ldots q_T|\Omega_1) = P(q_T|U(q_{[T:1]})|\Omega_1)\\ &= P(q_T|q_{T-1})P(q_{T-1}|\cdots|q_2)P(q_2|q_1)P(q_1|\Omega_1) = a_{q_{T-1},q_T}\cdots a_{q_1,q_2}\cdot \pi_{q_1}\end{aligned} \tag{2.1.9}$$

Fig (2.1.1) is the graphic representation of Eq. (2.1.9) for $T$ =3, starting from the right. The time evolution of the system P-ket for a given MSS now can be written as:

$$|\Omega_{Q_T}) \equiv U(q_{[T:1]})|\Omega_1) \underset{(1.4.2)}{=} |q_{T-1})P(q_{T-1}|q_{T-2})...P(q_2|q_1)P(q_1|\Omega_1) \tag{2.1.10}$$

Eq. (2.1.9) is the FJP for a VMM, and $P(q_T|\Omega_{Q_T})$ is identical to Eq. (1.3.4) or (1.4.6).

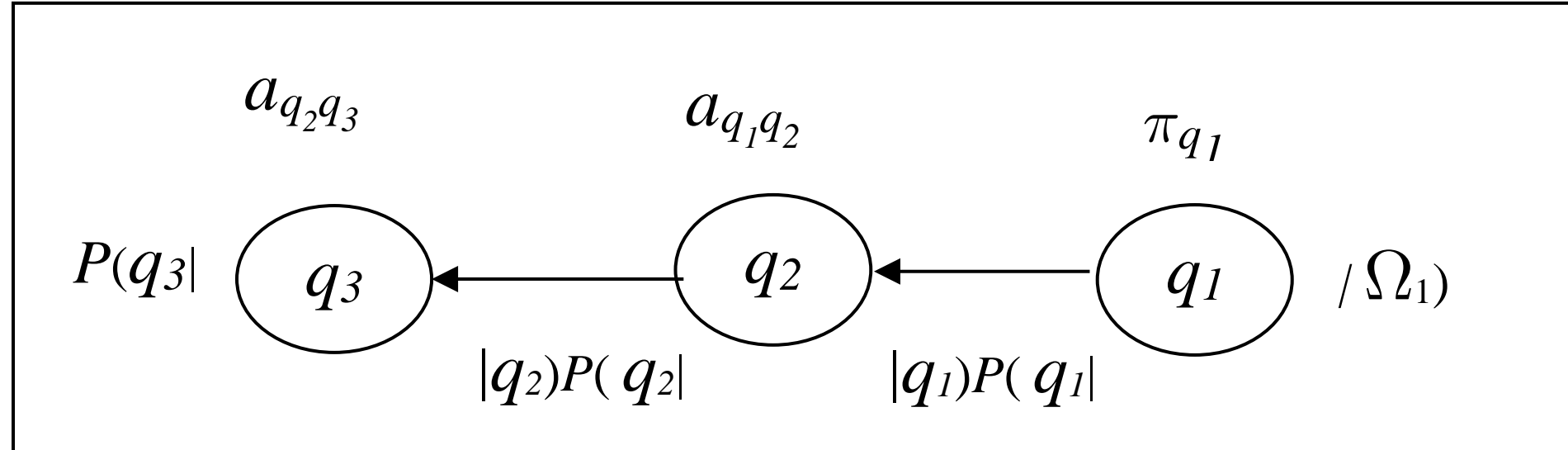

**Fig. 2.1.1**: The graphic representation of a VMM, Eq. (2.1.9) for $T$=3

## 2.2. The Fundamental VMM Formula

In a visible Markov model (VMM), a sequence of $T$ observations $O_T$ is conducted, and the observed value at each observation time is a state of the Markov state sequence:

$$O_T \equiv o_1 o_2 \cdots o_T \to Q_T \equiv Q_{[1:T]} \equiv \{q_1, q_2, \cdots, q_T\}, \quad q_t \in S, \quad 1 \le t \le T \tag{2.2.1}$$

What is the probability of the whole state sequence to happen? The answer is:

$$P(Q_T \mid \text{Model}) \equiv P(Q_{[1:T]} \mid \text{Model}) = P(q_1, q_2, \ldots q_T \mid \Lambda), \quad \Lambda \equiv \{\boldsymbol{A}, \boldsymbol{\pi}, \Omega\} \tag{2.2.2}$$

Here, we use the 3-parameter set $\Lambda$ to represent the model: the sample space Ω used in Eq. (2.1.9), the transition matrix $\boldsymbol{A}$ given by Eq. (1.2.3), and the initial probability distribution vector $\boldsymbol{\pi}$ given by Eq. (1.2.10). Note that only Ω is time-dependent, as described in Eq. (1.2.8). Thus, to denote the sample space Ω at time t, we can write:

$$\Lambda_t = \{\boldsymbol{A}, \boldsymbol{\pi}, \Omega_t\} \tag{2.2.3}$$

Now, our basic VMM formula, the full JP of a given state sequence, becomes

$$\begin{aligned} P(Q_T \mid \text{Model}) &= P(q_1, q_2, \ldots q_T \mid \Lambda_1) = P(q_T \mid U(q_{[T:1]}) \mid \Lambda_1) \\ &= a_{q_{T-1}, q_T} \cdots a_{q_1, q_2} \cdot \pi_{q_1} = \pi_{q_1} \cdot a_{q_1, q_2} \cdots a_{q_{T-1}, q_T} \end{aligned} \tag{2.2.4}$$

Here and from now on, we will use the simplified notation for the initial system P-ket:

$$|\Lambda_1) = |\{\boldsymbol{A}, \boldsymbol{\pi}, \Omega_1\}) \tag{2.2.5}$$

The basic formula is similar to [6, Eq. (4), pp. 2], except we do not include the state at time $t = 0$ ($q_0 \equiv q[0]$), and we have not yet set the initial state $\pi_{q_1}$. For a VMM, the model always starts with a fixed state [5, Eq. (4), pp. 259], that is:

$$q_1 = s_k \quad (1 \le k \le N) \quad \Rightarrow \quad \pi_j = \delta_{j,k} \tag{2.2.6}$$

Let us calculate the probability with the following observed state sequence [5] for our weather example:

$$Q_{[1:8]} = \{s_3, s_3, s_3, s_1, s_1, s_3, s_2, s_3\}, \quad \therefore \pi_j = \delta_{j,3} \tag{2.2.7}$$

From Eq. (2.2.4), we get:

$$P(s_3, s_3, s_3, s_1, s_1, s_3, s_2, s_3 \mid \Lambda_1) = a_{33} \cdot a_{33} \cdot a_{31} \cdot a_{11} \cdot a_{13} \cdot a_{32} \cdot a_{23} \cdot \pi_3 \tag{2.2.8}$$

Because $\pi_3 = 1$, Eq. (2.2.8) is identical to the answer given by [5, pp. 259].

### 2.3. The Sequential Event Space and its State P-Basis

**Sequential Event Space (SES):** The $T$ observations in Eq. (2.1.10) are conducted in a sequential event space at a given time sequence $t \in \{1, 2, \ldots T\}$:

$$\Omega \equiv \Omega_{[1:T]} = \Omega_1 \otimes \Omega_2 \otimes \ldots \otimes \Omega_T \tag{2.3.1}$$

The $P$-ket and $P$-bra for a particular state sequence $Q_T \equiv Q_{[1:T]}$ are given by:

$$P(Q_T| \equiv P(Q_{[1:T]}| \equiv P(q_1, q_2, \ldots q_T|, \quad |Q_T) \equiv |Q_{[1:T]}) \equiv |q_1, q_2, \ldots q_T) \tag{2.3.2}$$

**The P-Basis of an SES**: For a given time sequence, all possible Markov state sequences form a complete orthonormal set, as the P-basis of the sequential event space (SES) with the following symbolic properties:

Orthonormality: $P(Q_T|Q_T') \equiv P(q_1, q_2, \ldots q_T | q'_1, q'_2, \ldots q'_T) = \delta_{Q_T, Q_T'}$ (2.3.3)

Completeness: $I_{Q_T} = \sum_{all\ Q_T} |Q_T)P(Q_T| \equiv \sum_{all\ q_t} |q_1, q_1, q_2, \ldots q_T)P(q_1, q_2, \ldots q_T|$ (2.3.4)

Because there are $N$ possible states at any time $t$, the SES is $T^N$ -dimensional.

We can think of Eq. (2.3.3-4) as a natural extension of Eq. (1.1.1) (for a time-independent sample space) and Eq. (1.2.1) (for a sample space at a particular time $t$). The normalization condition of Eq. (1.1.3) now becomes:

$$1 = P(\Omega|\Omega) = P(\Omega|I_{Q_T}|\Omega) = P(\Omega|\sum_{Q_T}|Q_T)P(Q_T|\Omega) = \sum_{Q} P(Q_T|\Omega) \tag{2.3.5}$$

Here, we have used the following property of a conditional probability:

$$P(\Omega|Q_T) = 1, \quad \because Q_T \subseteq \Omega \tag{2.3.6}$$

For a given state sequence $Q_T$, Eq. (2.2.4) tells us the full JP:

$$P(Q_T|\Lambda_1) = P(q_1, q_2, \ldots q_T|\Lambda_1) = P(q_T|U(q_{[T:1]})|\Lambda_1) = a_{q_{T-1}, q_T} \cdots a_{q_1, q_2} \cdot \pi_{q_1} \tag{2.3.7}$$

The evolution of the system P-ket of the model can be symbolized as:

$$|\Lambda_{Q_T}) \equiv U(q_{[T:1]})|\Lambda_1) = |q_{T-1})\, a_{q_{T-2}, q_{T-1}} \cdots a_{q_1, q_2} \cdot \pi_{q_1} \tag{2.3.8}$$

The total probability for all possible state sequences is the sum of all possible FJP:

$$\sum_{all\ Q} P(Q_T|\Lambda_1) = \sum_{all\ q} P(q_1, q_2, \ldots q_T|\Lambda_1) = \sum_{all\ q} a_{q_{T-1}, q_T} \cdots a_{q_1, q_2} \cdot \pi_{q_1} \tag{2.3.9}$$

By comparing it with our expanded Markov evolution formula, Eq. (1.4.7), we have:

$$\sum_{all\ Q_T} P(Q_T \mid \Lambda_1) = \sum_{q_T} P(q_T \mid \Omega_T) = \sum_{q_T} P(q_T \mid \tilde{A}^{T-1} \mid \Omega_1) = 1 \qquad (2.3.10)$$

$$P(q_T \mid \tilde{A}^{T-1} \mid \Omega_1) = P(q_T \mid \Omega_T) = \sum_{Q_{T-1}} P(q_T \mid U(q_{[T:1]}) \mid \lambda_1) \qquad (2.3.11)$$

$$\mid \Omega_T) = \tilde{A}^{T-1} \mid \Omega_1) = \sum_{Q_{T-1}} U(q_{[T:1]}) \mid \lambda_1) \qquad (2.3.12)$$

Therefore, the evolution formula of a HMC, Eq. (1.4.7), is naturally associated with the FJP of a VMM, Eq. (2.3.7), which gives the joint probability of one specific Markov state sequence (MSS), $Q_{T-1} = \{q_1,\ldots,q_{T-1}\}$, among all possible paths of the HMC.

## 3. The PBN and Hidden Markov Models

In Visible Markov Models (VMMs), the state sequence $Q_T$ described in Eq. (2.2.1) is readily observable. In contrast, in Hidden Markov Models (HMMs) [3], the values we can observe are not the Markov states themselves but are derived from another random variable, denoted as $X$. A key task in HMMs is determining the most likely state sequence corresponding to a given sequence of observed values of $X_T$.

### 3.1. HMM and Probability Basis Transformation

In Problem 1.3.1 of our weather examples (§1.3), we are asked about the probability of a sequence of weather states. Imagine being in an isolated room where we cannot observe the weather directly. However, we can infer it by examining the surface of a magic stone. The stone may be wet or dry, depending on the weather conditions. This provides us with two detectable states to consider: wet or dry.

$$x[t] = x_t : \ t \rightarrow \omega_\mu \in W = \{\omega_1, \omega_2\} = \{\text{dry, wet}\} \qquad (3.1.1)$$

To make predictions by reasoning, we need the conditional probability of the state of the magic stone (whether dry or wet) given the weather conditions (sunny, foggy, or rainy) on the same day. Therefore, alongside the 3x3 matrix ***A*** presented in Equation (1.2.3), we have an additional 3x2 matrix ***B***:

$$\boldsymbol{B} = \{b_{i,\mu}\}: \quad b_{i,\mu} \equiv P(x_t = \omega_\mu \mid q_t = i), \quad 1 \le t \le T, \quad 1 \le i \le 3, \quad 1 \le \mu \le 2 \qquad (3.1.2)$$

The conditional probability matrix ***B*** relates to but quite differs from the unitary transformations of bases in Hilbert space, discussed in Quantum mechanics [7]. Suppose we have two observables, $S$ and $W$, in the Hilbert space. They both are Hermitian operators, but do not commute each other:

$$[\hat{S}, \hat{W}] \neq 0 \qquad (3.1.3)$$

Therefore, they don't share a set of common eigenvectors. Instead, each has a complete set of orthogonal unit vectors as a distinct basis within the same Hilbert space.

$$\hat{S}\,|\,s_i\rangle = s_i\,|\,s_i\rangle, \quad \langle s_i\,|\,s_k\rangle = \delta_{i,k}, \quad \sum_i |\,s_i\rangle\langle s_i\,| = I_S \tag{3.1.4a}$$

$$\hat{W}\,|\,\omega_\mu\rangle = \omega_\mu\,|\,\omega_\mu\rangle, \quad \langle \omega_\mu\,|\,\omega_\nu\rangle = \delta_{\mu,\nu}, \quad \sum_\mu |\,\omega_\mu\rangle\langle \omega_\mu\,| = I_W \tag{3.1.4b}$$

Given a system state $|\,\psi\rangle$, we can make the following unitary transformation:

$$\begin{aligned} |\,s_i\rangle &= I_W\,|\,s_i\rangle = \sum_\mu |\,\omega_\mu\rangle\langle \omega_\mu\,|\,s_i\rangle = \sum_i U_{\mu,i}\,|\,\omega_\mu\rangle \\ |\,\omega_\mu\rangle &= I_S\,|\,\omega_\mu\rangle = \sum_i |\,s_i\rangle\langle s_i\,|\,\omega_\mu\rangle = \sum_i U^\dagger{}_{i,\mu}\,|\,s_i\rangle \end{aligned} \tag{3.1.5}$$

Here, the unitary matrix $U$ is defined by:

$$U_{\mu,i} = \langle \omega_\mu\,|\,s_i\rangle, \quad U^\dagger U = U\,U^\dagger = I \tag{3.1.6}$$

The two bases in Eq. (3.1.4) in the Hilbert space can be adapted to probability space:

$$P(s_i\,|\,s_k) = \delta_{i,k}, \quad \sum_i |\,s_i)P(s_i\,| = I_S \tag{3.1.7a}$$

$$P(\omega_\mu\,|\,\omega_\nu) = \delta_{\mu,\nu}, \quad \sum_\mu |\,\omega_\mu)P(\omega_\mu\,| = I_W \tag{3.1.7b}$$

Similar to Eq. (3.1.6), we interpret the matrix $\boldsymbol{B}$ as the basis transformation in probability space at time = $t$, given by Eq. (3.1.2):

$$U_{\mu,i} = \langle \omega_\mu\,|\,s_i\rangle \rightarrow P(x_t = \omega_\mu\,|\,q_t = s_i) = P(\omega_\mu\,|\,s_i) \equiv b_{i,\mu}, \quad \boldsymbol{B} = \{b_{i,\mu}\} \tag{3.1.8a}$$

$$|\,s_i) = I_W\,|\,s_i) = \sum_\mu |\,\omega_\mu)P(\omega_\mu\,|\,s_i) = \sum_i b_{\mu,i}\,|\,\omega_\mu\rangle \tag{3.1.8b}$$

Suppose we know matrix $\boldsymbol{B}$ and the probability of $S$ at time $t$. The probability of $X$ at time $t$ can be calculated by inserting the unitary operator $I_S(t)$ into its $P$-bracket:

$$\begin{aligned} P(x_t = \omega_\mu\,|\,\Omega_t) &= P(\omega_\mu, t\,|\,I_S(t)\,|\,\Omega_t) = \sum_{i=1}^N P(\omega_\mu, t\,|\,s_i, t)P(s_i, t\,|\,\Omega_t) \\ &\equiv \sum_{i=1}^N b_{i,\mu} P(s_i, t\,|\,\Omega_t) \end{aligned} \tag{3.1.9}$$

Note that two important differences exist between the basis transformation in Hilbert and probability space. Firstly, the *matrix* $\boldsymbol{B}$ *is not a unitary matrix*. If we know the transform matrix in Eq. (3.1.8), then, to make its reverse transformation, we need to use the Bayesian formula:

$$P(s_i\,|\,\omega_\mu) = P(\omega_\mu\,|\,s_i)\frac{P(s_i)}{P(\omega_\mu)} = b_{i,\mu}\frac{P(s_i)}{P(\omega_\mu)} \tag{3.1.10}$$

Secondly, since $P(s_i, \omega_\mu) = P(s_i \mid \omega_\mu) P(\omega_\mu)$, *observables S and W can have a joint probability distribution* (or exact values simultaneously), they must be commuting observables in the probability space, which contrasts with the case in Hilbert space. For instance, the commutator of location $\hat{x}$ and momentum $\hat{p}$ in quantum mechanics leads to the famous Heisenberg uncertainty relation:

$$[\hat{x}, \hat{p}] = i\hbar \quad \Rightarrow \quad \Delta x \, \Delta p \geq \frac{\hbar}{2} \tag{3.1.11}$$

Thus, the observables ($\hat{x}$ and $\hat{p}$) cannot have exact values simultaneously, and the following expressions are illegal (meaningless) in quantum probability modeling:

$$P(x, p), \quad P(x \mid p), \quad P(p \mid x)... \tag{3.1.12}$$

### 3.2. The Fundamental HMM Formulas and the Observed P-Basis

The question at hand is: Given a sequence of observed conditions of stones in the room, such as {wet, wet, dry, wet, dry}, what are the most likely sequences of weather states outside the building? To answer this question, we must understand some fundamental Hidden Markov Models (HMM) concepts.

We denote the sequence of observations $O_T$ and the associated sequence of observed values of $X$ as:

$$O_T \equiv o_1 o_2 \cdots o_T \rightarrow X_T \equiv X_{[1:T]} = \{x_1, x_2, \cdots x_T\} \tag{3.2.1}$$

The observed values of $X$ are in the set of $W$ with $K$ elements:

$$x[t] = x_t : \, t \rightarrow \omega_\mu \in W = \{\omega_1, \omega_2, \ldots \omega_K\}, \quad 1 \leq \mu \leq K \tag{3.2.2a}$$

To avoid confusion, we will use Greek letters $(\mu, \nu, \ldots)$ to label the value in sequence $X$, and Latin letters $(i, \, j, \ldots)$ to label states of the Markov state sequence.

The observed set $W$ constructs the basis of a vector space as shown in Eq. (3.1.4):

Orthogonality: $\langle \mu \mid \nu \rangle \equiv \langle \omega_\mu \mid \omega_\nu \rangle = \delta_{\mu\nu}$ (3.2.2b)

Completeness: $\sum_{\mu=1}^{K} \mid \mu \rangle \langle \mu \mid \equiv \sum_{\mu=1}^{K} \mid \nu_\mu \rangle \langle \nu_\mu \mid = I_W$ (3.2.2c)

They can be mapped to the $P$-basis of the observed variable $X$:

Orthogonality: $P(\mu \mid \nu) \equiv P(\omega_\mu \mid \omega_\nu) = \delta_{\mu\nu}$ (3.2.2d)

Completeness: $\sum_{\mu=1}^{K} \mid \omega_\mu ) P(\omega_\mu \mid = I_W$ (3.2.2e)

The conditional probability of $x_t = v_\mu$ given $q_t = s_i$ forms an observation-state matrix $\boldsymbol{B}$:

$$\boldsymbol{B} = \{b_{i,\mu}\}: \quad b_{i,\mu} \equiv b_i(\omega_\mu) = P(x_t = \omega_\mu \mid q_t = i), \quad 1 \le t \le T, \quad 1 \le i \le N, \quad 1 \le \mu \le K \quad (3.2.3a)$$

Matrix $\boldsymbol{B}$ is our transformation matrix of the two P-bases, defined in Eq. (3.1.8). Note that $\boldsymbol{B}$ is assumed time-independent, and it satisfies the following requirement:

$$\sum_{\mu=1}^{K} b_{i,\mu} = \sum_{\mu=1}^{K} b_i(\omega_\mu) = \sum_{\mu=1}^{K} P(\omega_\mu \mid i) = 1, \qquad 1 \le i \le N \quad (3.2.3b)$$

The requirement can be easily derived by using the PBN:

$$\begin{aligned} 1 &\underset{(1.1.5)}{=} P(W \mid i) = P(W \mid I_W \mid i) \underset{(2.1.2)}{=} P(W \mid \left\{ \sum_{\mu=1}^{K} \mid \omega_\mu) P(\omega_\mu \mid \right\} \mid i) \\ &= \sum_{\mu=1}^{K} b_i(\omega_\mu) P(W \mid \omega_\mu) \underset{(1.1.5)}{=} \sum_{\mu=1}^{K} b_i(\omega_\mu) = 1 \end{aligned} \quad (3.2.3c)$$

The HMM model $\Xi$ is defined by the set of four parameters: the transition matrix $\boldsymbol{A}$, observation-state matrix $\boldsymbol{B}$, initial state distribution vector $\boldsymbol{\pi}$ and $\Omega$:

$$\Xi = \{\boldsymbol{B}, \boldsymbol{A}, \boldsymbol{\pi}, \Omega\} = \{\boldsymbol{B}, \Lambda\} \quad (3.2.4a)$$

Note that only $\Omega$ is time-dependent, as described in Eq. (1.2.8). Therefore, to denote the sample space $\Omega$ at time t, we can write:

$$\Xi_t = \{\boldsymbol{B}, \boldsymbol{A}, \boldsymbol{\pi}, \Omega_t\} = \{\boldsymbol{B}, \Lambda_t\} \quad (3.2.4b)$$

Four basic formulas exist for an HMM [5] [6].

**HMM F-1.** The FJP of MSS $Q_T$: It is identical to Eq. (2.2.4) for a VMM:

$$\begin{aligned} P(Q_T \mid \Xi_1) &= P(q_1, q_2, \ldots q_T \mid \Xi_1) = \prod_{t=1}^{T-1} P(q_{t+1} \mid q_t) \cdot \pi_{q_1} \\ &= \prod_{t=1}^{T-1} a_{q_t, q_{t+1}} \cdot \pi_{q_1} = \pi_{q_1} \cdot a_{q_1, q_2} \cdot a_{q_2, q_3} \cdots a_{q_{T-1}, q_T} \end{aligned} \quad (3.2.5a)$$

Like Eq. (2.2.5) for VMM, we use the following system P-ket for a HMM:

$$|\Xi_1) = |\{\boldsymbol{B}, \boldsymbol{A}, \boldsymbol{\pi}, \Omega_1\}) \quad (3.2.5b)$$

Note that Eq. (3.2.5a) equals Eq. (2.1.9) if replacing $|\Xi_1)$ with $|\Omega_1)$, and there is a one-to-one map between the initial sample space $|\Omega_1)$ and $\boldsymbol{\pi}$:

$$|\Omega_1) = \sum_{i=1}^{N} \pi_i \mid s_i, t=1), \quad \pi_i = P(s_i, t=1 \mid \Omega_1) \quad (3.2.5c)$$

**HMM F-2.** Assuming that such observations are independent of each other, we can use Eq. (3.2.3) to write the conditional probability of an observed sequence $X_{[1\to T]}$ given an MSS $Q_{[1:T]}$ as:

$$
\begin{aligned}
&P(X \mid Q) = P(X_{[1:T]} \mid Q_{[1:T]}) = P(x_1, x_2, \dots x_T \mid q_1, q_2, \dots q_T) \\
&= \prod\nolimits_{t=1}^{T} P(x_t \mid q_t) \underset{(2.1.3)}{=} \prod\nolimits_{t=1}^{T} b_{q_t}(x_t) = b_{q_1,\, x_1} \cdot b_{q_2,\, x_2} \cdots b_{q_N,\, x_N}
\end{aligned} \tag{3.2.6}
$$

Eq. (3.2.6) can be interpreted as the extension of the unitary transformation, given by Eq. (3.1.6), to the P-basis transformation from the P-basi**s** in the $T^N$ -dimensional SES for the hidden MSS $Q_T$, as described by Eq. (2.3.3), to the observed P-basis in the $T^K$ dimensional SES for visible $X_T$, which has the following properties:

Orthonormality: $P(X \mid X') \equiv P(x_1, x_2, \dots x_T \mid x'_1, x'_2, \dots x'_T) = \delta_{X,X'}$ (3.2.7a)

Completeness: $I_X = \sum\nolimits_{all\ X} |\, X) P(X \,| \equiv \sum\nolimits_{all\ X} |\, x_1, x_2, \dots x_T) P(x_1, x_2, \dots x_T \,|$ (3.2.7b)

**HMM F-3.** The joint likelihood (joint probability) of an observation sequence *X* [1:*T*] and a state sequence *Q* [1:*T*]. Using the Bayes theorem, it reads:

$$
P(X_T, Q_T \mid \Xi_1) = P(X_T \mid Q_T) \cdot P(Q_T \mid \Xi_1) \tag{3.2.7}
$$

It is the full JP in the SES of the HMM. Using Eq. (3.2.5-6), it can be expanded as:

$$
\begin{aligned}
&P(X_T, Q_T \mid \Xi_1) = P(q_1, \dots q_T; x_1, \dots, x_T \mid \Xi_1) = P(X_T \mid Q_T) \cdot P(Q_T \mid \Xi_1) \\
&= P(x_1, \dots, x_T \mid q_1, \dots q_T) \cdot P(q_1, \dots q_T \mid \Xi_1) = b_{q_1,\, x_1} \cdot a_{q_1,\, q_2} \cdot b_{q_2,\, x_2} \cdots a_{q_{T-1},\, q_T} \cdot b_{q_T,\, x_T} \cdot \pi_{q_1}
\end{aligned} \tag{3.2.8}
$$

**VMM F-4**. The marginal probability of observation sequence *X*[1:*T*]: It is the sum of the joint probability of sequence $X_T$ with all possible state sequences *Q*[1:*T*]:

$$
P(X_T \mid \Xi_1) = \sum\nolimits_{all\ Q_T} P(X_T, Q_T \mid \Xi_1) \tag{3.2.9}
$$

It can be expanded by using Eq. (3.2.8):

$$
\begin{aligned}
&P(X_T \mid \Xi_1) = \sum\nolimits_{all\ Q} P(X_T, Q_T \mid \Xi_1) = \sum\nolimits_{all\ Q_T} P(X_T \mid Q_T) P(Q_T \mid \Xi_1) \\
&= \sum\nolimits_{all\ q} \pi_{q_1} \cdot b_{q_1,\, x_1} \cdot a_{q_1,\, q_2} \cdot b_{q_2,\, x_2} \cdots a_{q_{T-1},\, q_T} \cdot b_{q_T,\, x_T}
\end{aligned} \tag{3.2.10}
$$

Because Eq. (3.2.10) is valid for any *X*, we obtain an identity operator in the sequential event space for hidden variable *Q*, as already shown in Eq. (2.3.4):

$$I_{Q_T} = \sum\nolimits_{all\ Q_T} |\, Q_T)\, P(Q_T \,| = \sum\nolimits_{all\ q} |\, q_1, q_2, ...q_T)\, P(q_1, q_2, ...q_T \,| \qquad (3.2.11)$$

Using this identity operator, Eq. (3.2.10) can be re-derived as:

$$\begin{aligned} P(X_T \,|\, \Xi_1) &= P(X_T \,|\, I_{Q_T} \,|\, \Xi_1) \underset{(3.2.11)}{=} \sum\nolimits_{all\ Q} P(X_T \,|\, Q_T) P(Q_T \,|\, \Xi_1) \\ &\underset{(3.2.10)}{=} \sum\nolimits_{all\ q} \pi_{q_1} \cdot b_{q_1,\,x_1} \cdot a_{q_1,\,q_2} \cdot b_{q_2,\,x_2} \cdots a_{q_{T-1},\,q_T} \cdot b_{q_T,\,x_T} \end{aligned} \qquad (3.2.12)$$

Eq. (3.2.10) or (3.2.12) requires around $2T \cdot N^T$ calculations. It is unrealistic to conduct such calculations even for small values like $N = 5$ and $T = 100$ [5]. Fortunately, there are recursive algorithms to avoid such a problem. For example, the most likely state sequence given an observed sequence can be obtained using the Viterbi algorithm [6, 8].

Eq. (3.2.5a), the time evolution of a hidden Markov sequence, can be symbolized as:

$$|\, \Xi_{Q_T}) \equiv U_{[1:T]} \,|\, \Xi_1) \underset{(3.3.7)}{=} \sum\nolimits_{all\ q} |\, q_T) \pi_{q_1} \cdot b_{q_1,\,x_1} \cdot a_{q_1,\,q_2} \cdot b_{q_2,\,x_2} \cdots a_{q_{T-1},\,q_T} \qquad (3.2.13)$$

$$\begin{aligned} P(q_T \,|\, \Xi_{Q_T}) &= P(q_T \,|\, U_{[1:T]} \,|\, \Xi_1) = P(q_1, q_2, \ldots q_T \,|\, \Xi_1) = \prod\nolimits_{t=1}^{T-1} P(q_{t+1} \,|\, q_t) \cdot \pi_{q_1} \\ &= \prod\nolimits_{t=1}^{T-1} a_{q_t,\, q_{t+1}} \cdot \pi_{q_1} = \pi_{q_1} \cdot a_{q_1,\,q_2} \cdot a_{q_2,\,q_3} \cdots a_{q_{T-1},\,q_T} \end{aligned} \qquad (3.2.14)$$

### 3.3. The Optimal State Sequence and the Viterbi algorithm

In speech recognition and several other pattern recognition applications [9], it is useful to associate an “optimal” sequence of states to a sequence of observations, given the parameters of a model. For instance, in speech recognition, knowing which frames of features “belong” to which state allows one to locate the word boundaries across time.

A “reasonable” optimality criterion consists of choosing $Q_T$, the state sequence (or path), having the maximum likelihood for a given observed sequence $X_T$ of the model; the hidden sequence $Q_T$ can be determined recursively via the Viterbi algorithm [6]. This algorithm makes use of two variables (with our notation):

- $\delta_t(i)$ is the highest likelihood of a single path among all the paths ending in the state $s_i = i$ at time $t$ (i.e., $q_t = i$):
  $$\delta_t(i) = \delta_t(q_t = i) = \max_{q_1, q_2, \ldots q_{t-1}} P(q_1, q_2, \ldots, q_{t-1}, q_t = i; x_1, x_2, \ldots, x_t \,|\, \lambda_1) \qquad (3.3.1)$$
- a variable $\psi_t(i)$, which allows to keep track of the “best path” ending in the state $s_i = i$ at time $t$ (i.e., $q_t = i$):
  $$\psi_t(i) = \arg\max_{q_1, q_2, \ldots q_{t-1}} P(q_1, q_2, \ldots, q_{t-1}, q_t = s_i; x_1, x_2, \ldots, x_t \,|\, \lambda_1) \qquad (3.3.2)$$

Using Eq. (3.2.8), we can derive a recursive relation for $\delta_t(i)$:

$$\delta_t(j) = \delta_t(q_t = j) = \max_{q_1, q_2, \ldots q_{t-1}} \left\{ b_{q_1, x_1} \cdot a_{q_1, q_2} \cdot b_{q_2, x_2} \cdots a_{q_{t-1}, q_t = j} \cdot b_{q_t = j, x_t} \cdot \pi_{q_1} \right\}$$

$$\delta_{t-1}(i) = \delta_{t-1}(q_{t-1} = i) = \max_{q_1, q_2, \ldots q_{t-2}} \left\{ b_{q_1, x_1} \cdot a_{q_1, q_2} \cdot b_{q_2, x_2} \cdots a_{q_{t-2}, q_{t-1} = i} \cdot b_{q_{t-1} = i, x_{t-1}} \cdot \pi_{q_1} \right\}$$

$$\begin{aligned} \therefore\ \delta_t(j) &= \delta_t(q_t = j) = \max_{q_{t-1} = i} \left\{ \delta_{t-1}(i) \cdot a_{q_{t-1} = i, q_t = j} \cdot b_{q_t = j, x_t} \right\} \\ &= \max_{1 \le i \le N} \left\{ \delta_{t-1}(i) \cdot a_{i, j} \right\} \cdot b_{j, x_t} \end{aligned} \tag{3.3.3}$$

Similarly, by using Eq. (3.2.8), we can derive a recursive relation for $\psi_t(i)$:

$$\psi_t(j) = \psi_t(q_t = j) = \arg\max_{q_1, q_2, \ldots q_{t-1}} \left\{ b_{q_1, x_1} \cdot a_{q_1, q_2} \cdot b_{q_2, x_2} \cdots a_{q_{t-1}, q_t = j} \cdot b_{q_t = j, x_t} \cdot \pi_{q_1} \right\}$$

$$\psi_{t-1}(i) = \psi_{t-1}(q_{t-1} = i) = \arg\max_{q_1, q_2, \ldots q_{t-2}} \left\{ b_{q_1, x_1} \cdot a_{q_1, q_2} \cdot b_{q_2, x_2} \cdots a_{q_{t-2}, q_{t-1} = i} \cdot b_{q_{t-1} = i, x_{t-1}} \cdot \pi_{q_1} \right\}$$

$$\begin{aligned} \therefore\ \psi_t(j) &= \psi_t(q_t = j) = \arg\max_{q_{t-1} = i} \left\{ \delta_{t-1}(q_{t-1} = i) \cdot a_{q_{t-1} = i, q_t = j} \cdot b_{j, x_t} \right\} \\ &= \arg\max_{1 \le i \le N} \left\{ \delta_{t-1}(i) \cdot a_{i, j} \right\} \quad \text{(Note: } b_{j, x_t} \text{ has no effect on arg max)} \end{aligned} \tag{3.3.4}$$

Now we list all the formulas required for the Viterbi algorithm [6] using our notation:

1. **Initialization:**

$$\begin{aligned} &\delta_1(i) = \pi_i \cdot b_{i, x_1}, \quad i = 1, \ldots, N \\ &\psi_1(i) = 0 \end{aligned} \tag{3.3.5}$$

   Here $\pi_i$ is the prior probability of being in the state $s_i$ at time $t = 1$.

2. **Recursion:**

$$\begin{aligned} &\delta_t(j) = \max_{1 \le i \le N} \left\{ \delta_{t-1}(i) \cdot a_{i, j} \right\} \cdot b_{j, x_t}, \quad 2 \le t \le T,\ 1 \le j \le N \\ &\psi_t(j) = \arg\max_{1 \le i \le N} \left\{ \delta_{t-1}(i) \cdot a_{i, j} \right\}, \quad 2 \le t \le T,\ 1 \le j \le N \end{aligned} \tag{3.3.7}$$

   Therefore, $\psi_t(j)$ stores the most probable state at time $= t - 1$ given $s_t = j$.

3. **Termination:**

$$\begin{aligned} &P^*(X \mid \lambda) = \max_{1 \le i \le N} \delta_T(i) \\ &q_T^* = \arg\max_{1 \le i \le N} \delta_T(i) \end{aligned} \tag{3.3.8}$$

   At the termination, the best likelihood state is found when the end of the observation sequence $t = T$ is reached.

4. **Backtracking:**

$$Q_T^* = \{q_1^*, \ldots, q_T^*\} \quad \text{so that} \quad q_t^* = \psi_{t+1}(q_{t+1}^*), \quad t = T-1,\ T-2, \ldots, 1 \tag{3.3.9}$$

Starting from $q_T^*$ in Eq. (3.3.8), one decodes the most likely state at the previous time, $q_{T-1}^* = \psi_T(q_T^*)$ and so on. In the end, the best sequence of states $Q_T$* is obtained.

# 4. The Weather-Stone Example: Viterbi Algorithm

This section will show the details of applying the Viterbi algorithm to the Weather-Magic Stone example adapted from [6] with our notation. Then, we will use Elvira, a Java software package, to graphically display the example as a case study of probabilistic graphic models [10], including dynamic Bayesian Networks[10, 11].

## 4.1. The Weather-Stone Example and the Viterbi Algorithm

How does the Viterbi algorithm work? Let's revisit the weather-magic stone example. In this scenario, weather and the magic stone represent two commuting observables. The weather consists of three states and has a transition matrix A, defined by Eq. (1.3.1). The sequence of weather states *Q* is hidden from the observer. The magic stone has two detectable values, as mentioned in Eq. (3.1.1), and we know its sequence *X*. How can we determine the most likely state sequence *Q* given the observed sequence *X*?

First, we need to define the ***B*** matrix of Eq. (3.2.3) or PBT of Eq. (3.1.6). We will use the same data as in [6] but with a magic stone instead of an umbrella:

| Weather (Given state) | Probability of Dry Stone ($\mu = 1$, or no umbrella) | Probability of Wet Stone ($\mu = 2$, or with umbrella) |
|---|---|---|
| Sunny ( i = 1) | 0.9 | 0.1 |
| Rainy ( i = 2) | 0.2 | 0.8 |
| Foggy ( i = 3) | 0.7 | 0.3 |

**Table 3.4.1**: The Weather-Stone (or Umbrella) Correlation

From the data, we obtain the following ***B*** matrix:

$$\boldsymbol{B} = \{b_{i,\mu}\} = \{P(x_t = \mu \mid q_t = i)\} = \begin{pmatrix} 0.9 & 0.1 \\ 0.2 & 0.8 \\ 0.7 & 0.3 \end{pmatrix} \tag{4.1.1}$$

**Task:** You don't know how the weather was while locked in the isolated room. On the first 3 days, your stone observations are:

$$X = \{x_1, x_2, x_3\} = \{1,2,2\} = \{\text{dry,wet,wet}\} \tag{4.1.2a}$$

How do we use the Viterbi algorithm in Eqs (3.3.5-9) to find the most probable weather sequence? We assume the 3 weather situations to be equal-probable on day 1:

$$\pi_i = 1/3, \quad i \in \{1,2,3\} \ . \tag{4.1.2b}$$

There is nothing new than what is stated in [6]. We repeat the calculation here by using our notation (also replacing small icons in [6] with the words or labels they represent), hoping it is more convenient for our readers.

1. **Initialization** (day 1: $t$ = 1):

$i = 1$ (sunny):

$$\delta_{t=1}(\text{sunny}) = \delta_1(1) = \pi_1 \cdot b_{1,\,x_1} = \pi_1 \cdot b_{1,1} = \frac{1}{3} \cdot b_{\text{sunny, dry}} = \frac{1}{3} \cdot 0.9 = 0.3 \tag{4.1.3a}$$

$$\psi_{t=1}(\text{sunny}) = \psi_1(1) = 0$$

$$\delta_{t=1}(\text{rainy}) = \delta_1(2) = \pi_2 \cdot b_{2,1} = \frac{1}{3} \cdot b_{\text{rainy, dry}} = \frac{1}{3} \cdot 0.2 = 0.0667 \tag{4.1.3b}$$

$$\psi_{t=1}(\text{rainy}) = \psi_1(2) = 0$$

$$\delta_{t=1}(\text{foggy}) = \delta_1(3) = \pi_3 \cdot b_{3,\,x_1} = \pi_3 \cdot b_{3,1} = \frac{1}{3} \cdot b_{\text{foggy, dry}} = \frac{1}{3} \cdot 0.7 = 0.233 \tag{4.1.3c}$$

$$\psi_{t=1}(\text{foggy}) = \psi_1(3) = 0$$

2. **Recursion**:

**Day 2: *t* = 2**

$i = 1$ (sunny), $\mu = 2$(wet):

$$\begin{aligned} \delta_{t=2}(\text{sunny}) &= \max\{\delta_1(1) \cdot a_{1,1},\, \delta_1(2) \cdot a_{2,1},\, \delta_1(3) \cdot a_{3,1}\} \cdot b_{1,2} \\ &= \max\{\mathbf{0.3 \times 0.8},\, 0.0667 \cdot 0.2,\, 0.233 \cdot 0.2\} \cdot 0.1 = 0.024 \end{aligned} \tag{4.1.4a}$$

$$\therefore \psi_{t=2}(\text{sunny}) = \text{ sunny} = s_1$$

$i = 2$ (rainy), $\mu = 2$(wet):

$$\begin{aligned} \delta_{t=2}(\text{rainy}) &= \max\{\delta_1(1) \cdot a_{1,2},\, \delta_1(2) \cdot a_{2,2},\, \delta_1(3) \cdot a_{3,2}\} \cdot b_{2,2} \\ &= \max\{0.3 \cdot 0.05,\, 0.0667 \times 0.6,\, \mathbf{0.233 \times 0.5}\} \cdot 0.8 = 0.056 \end{aligned} \tag{4.1.4b}$$

$$\therefore \psi_{t=2}(\text{rainy}) = \text{ foggy} = s_3$$

$i = 3$ (foggy), $\mu = 2$(wet):

$$\begin{aligned} \delta_{t=2}(\text{foggy}) &= \max\{\delta_1(1) \cdot a_{1,3},\, \delta_1(2) \cdot a_{2,3},\, \delta_1(3) \cdot a_{3,3}\} \cdot b_{3,2} \\ &= \max\{0.3 \cdot 0.15,\, 0.0667 \cdot 0.2,\, \mathbf{0.233 \times 0.5}\} \cdot 0.3 = 0.035 \end{aligned} \tag{4.1.4c}$$

$$\therefore \psi_{t=2}(\text{foggy}) = \text{ foggy} = s_3$$

**Day 3: *t* = 3**

$$
\begin{aligned}
& i = 1 \text{ (sunny)}, \quad \mu = 2 \text{ (wet)}: \\
& \delta_{t=3}(\text{sunny}) = \max\{\delta_2(1)\cdot a_{1,1},\, \delta_2(2)\cdot a_{2,1},\, \delta_2(3)\cdot a_{3,1}\}\cdot b_{1,2} \\
& \qquad = \max\{\mathbf{0.024}\times\mathbf{0.8},\, 0.056\cdot 0.2,\, 0.035\cdot 0.2\,\}\cdot 0.1 \\
& \qquad = \max\{\mathbf{0.019},\, 0.011,\, 0.007\,\}\cdot 0.1 = 0.0019 \\
& \therefore \psi_{t=3}(\text{sunny}) = \text{ sunny} = s_1
\end{aligned}
\tag{4.1.5a}
$$

$$
\begin{aligned}
& i = 2 \text{ (rainy)}, \quad \mu = 2\text{(wet)}: \\
& \delta_{t=3}(\text{rainy}) = \max\{\delta_2(1)\cdot a_{1,2},\, \delta_2(2)\cdot a_{2,2},\, \delta_2(3)\cdot a_{3,2}\}\cdot b_{2,2} \\
& \qquad = \max\{0.024\cdot 0.05,\, \mathbf{0.056}\times\mathbf{0.6},\, 0.035\times 0.3\,\}\cdot 0.8 \\
& \qquad = \max\{0.0012,\, \mathbf{0.0336},\, 0.00105\,\}\cdot 0.8 = 0.0269 \\
& \therefore \psi_{t=3}(\text{rainy}) = \text{ rainy} = s_2
\end{aligned}
\tag{4.1.5b}
$$

$$
\begin{aligned}
& i = 3 \text{ (foggy)}, \quad \mu = 2\text{(wet)}: \\
& \delta_{t=3}(\text{foggy}) = \max\{\delta_2(1)\cdot a_{1,3},\, \delta_2(2)\cdot a_{2,3},\, \delta_2(3)\cdot a_{3,3}\}\cdot b_{3,2} \\
& \qquad = \max\{0.0024\cdot 0.15,\, 0.056\cdot 0.2,\, \mathbf{0.035}\times\mathbf{0.5}\,\}\cdot 0.3 \\
& \qquad = \max\{0.00036,\, 0.0112,\, \mathbf{0.0178}\,\}\cdot 0.3 = 0.0052 \\
& \therefore \psi_{t=3}(\text{foggy}) = \text{ foggy} = s_3
\end{aligned}
\tag{4.1.5c}
$$

3. **Termination**

$$
\begin{aligned}
& P^*(X \mid \lambda) = \max_{1\le i\le 3} \delta_3(i) = \delta_3(\text{rainy}) = 0.0269 \\
& q_3^* = \arg\max_{1\le i\le 3} \delta_3(i) = \text{rainy} = s_2
\end{aligned}
\tag{4.1.6}
$$

4. **Backtracking**

$$
t = 2: \quad q_2^* = \psi_3(q_3^*) = \psi_3(\text{rainy}) = \text{rainy} = s_2 \tag{4.1.7a}
$$

$$
t = 1: \quad q_1^* = \psi_2(q_2^*) = \psi_2(\text{rainy}) = \text{foggy} = s_3 \tag{4.1.7b}
$$

Therefore, the **most likely weather sequence** is:

$$
Q^* = \{q_1^*, q_2^*, q_3^*\} = \{\text{foggy, rainy, rainy}\} = \{s_3, s_2, s_2\} \tag{4.1.8}
$$

### 4.2. The Weather-Stone HMM Example and the Elvira Software

There are numerous software packages [13] to manipulate static or dynamic Bayesian networks. Elvira [14] is one of them[3]. It is written in Java and has a lovely user graphic interface. In Ref. [12], we have used it to display the static student BN [10]. Now we apply it to our Weather-Stone HMM model with the initial state distribution in Eq. (4.1.2b), the transition matrix as in Eq. (1.3.1), the transformation matrix as in Eq. (4.1.1)

---

[3] The Elvira software is not available online anymore. One may try tools such as the Bayes Server [23].

and set the observed values as in Eq. (4.1.2a). The screenshot of the probabilistic graphic model (PGM) in Edit mode is given in Figure (4.2.1) below.

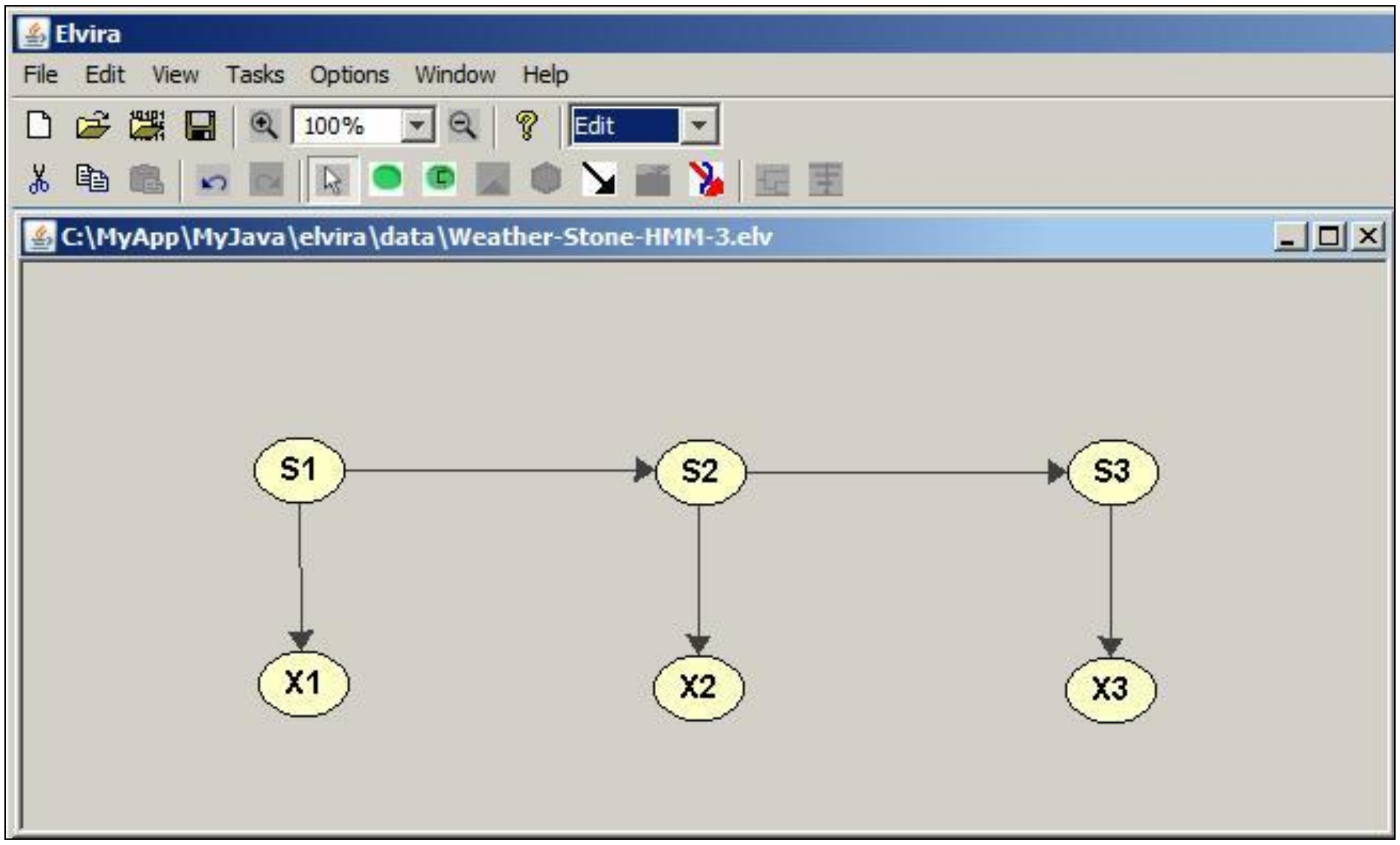

**Fig. 4.2.1**: The Weather-Stone HMM in Elvira Edit mode.

Then we change to Elvira Inference mode. Without setting any observed values of the magic stone, we have the prior probability distributions shown in Figure (4.2.2).

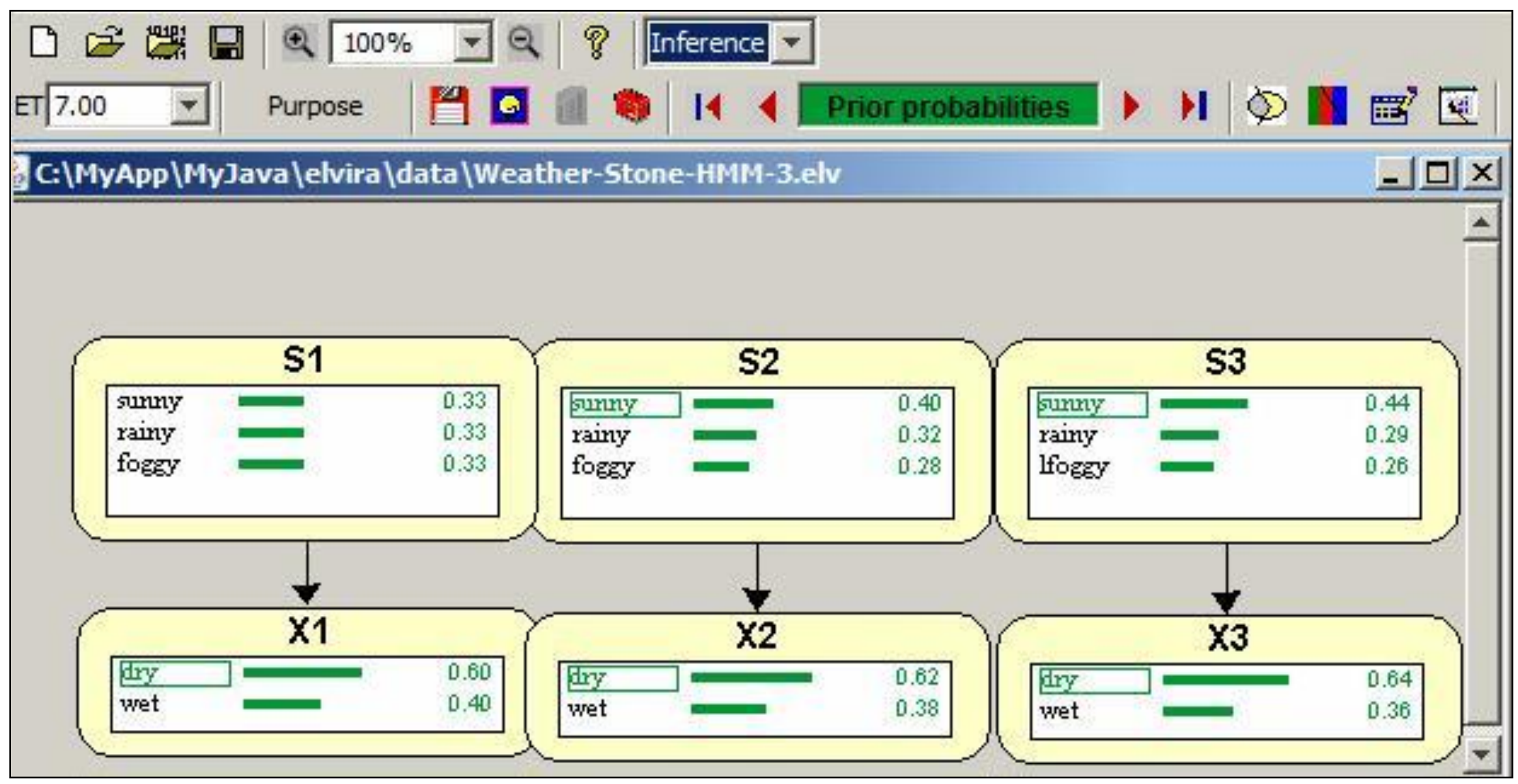

**Fig. 4.2.2**: The Prior Properties of Weather-Stone HMM in Elvira Inference mode.

Then, in Elvira Inference mode, we set the observed stone values for *X1*, *X2,* and *X3* as in (4.1.2). The screenshot is shown in Figure (4.2.3): given the observed stone value sequence {dry, wet, wet}, the most likely weather state sequence is {foggy, rainy, rainy}, consistent with our numerical result using the Viterbi algorithm, Eq. (4.1.8).

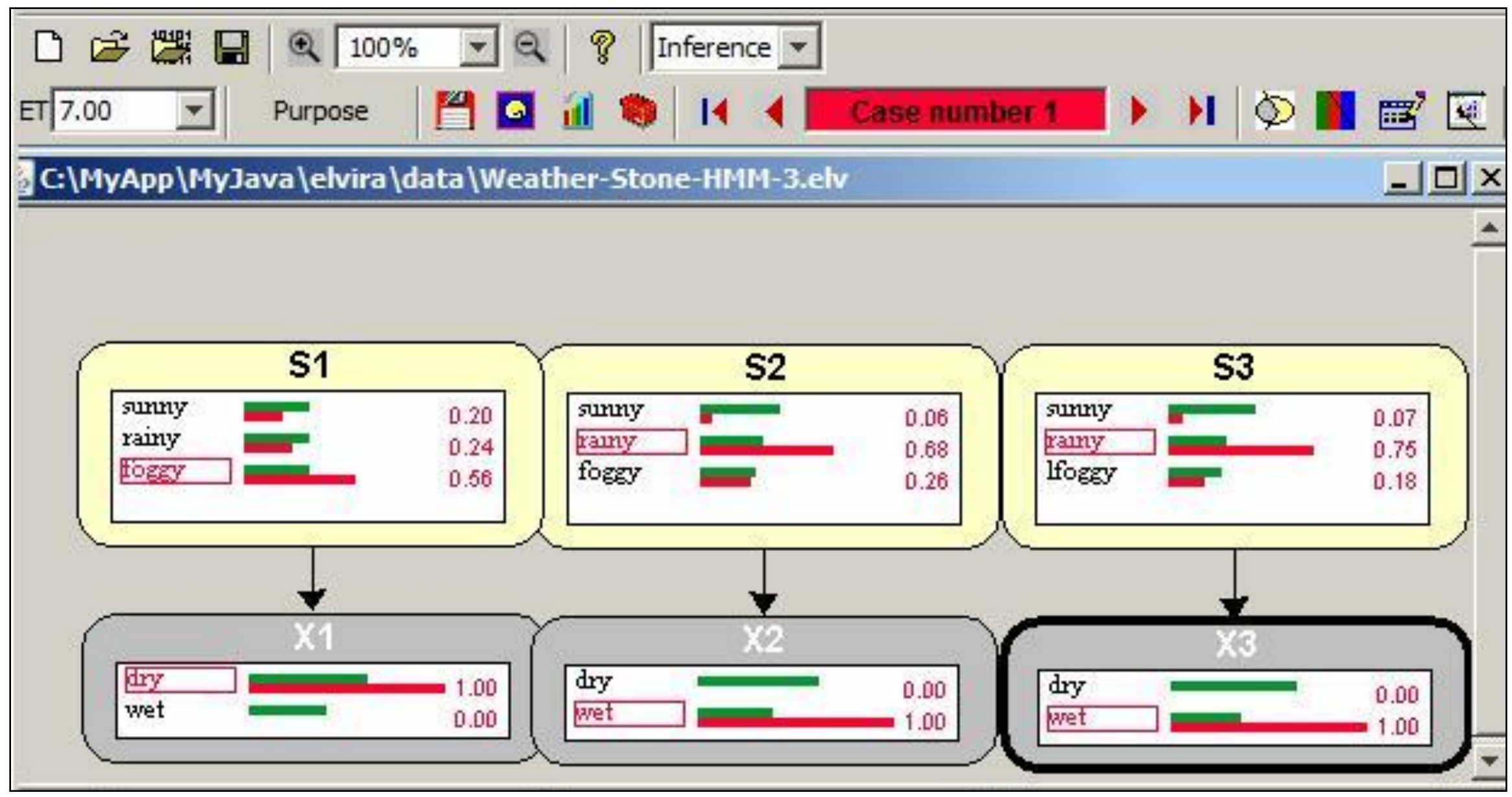

**Fig. 4.2.3**: The Weather-Stone HMM with explicit observed stone values.

The related Elvira code file is available online [15].

## 5. The VMM, HMM, and FHMM as Dynamic Bayesian Networks

The VMM, HMM, and Factorial HMM (FHMM) are systems of multiple random variables with various probability distributions. They are classified as Dynamic Bayesian Networks (DBN) within Probabilistic Graphic Models (PGM) [10].

### 5.1. The VMM as a DBN of *T*-nodes

If we take the *T* states in the sequential event space in Eq. (2.2.1) as a *T*-node DBN, then its graph in PGM is shown in Fig. 5.1.1 below.

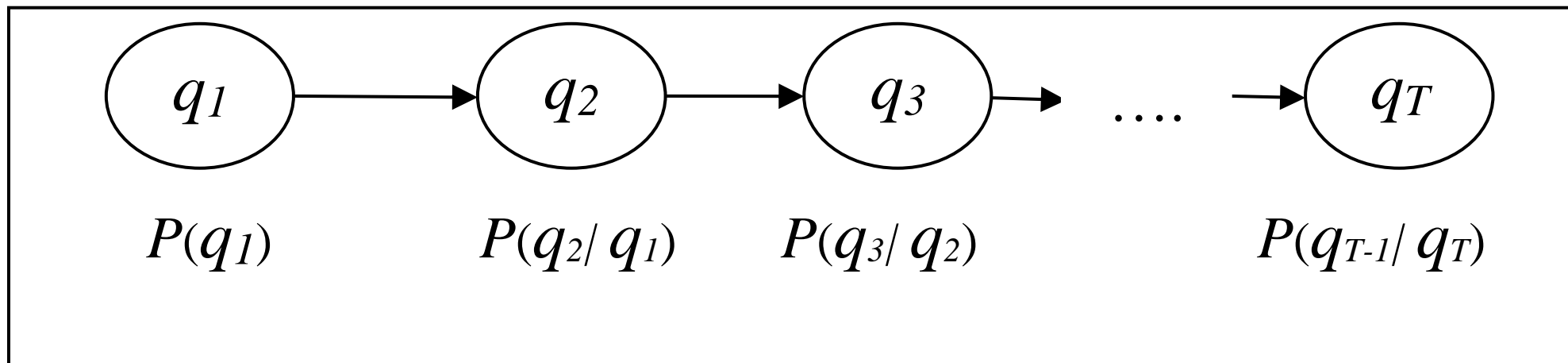

**Fig. 5.1.1**: A Visible Markov Model as a Dynamic Bayesian Network

Reading from the graph, we have the following full JP:

$$P(Q_T) \equiv P(q_1,\ldots,q_T) \;= P(q_1)P(q_2 \mid q_1)\cdots P(q_T \mid q_{T-1}) \tag{5.1.1}$$

Applying Eq. (1.2.5) and (1.2.10), we get:

$$P(Q_T) = P(q_1,\ldots,q_T) \;= \pi_{q_1} a_{q_1,q_2} \cdots a_{q_{T-1},q_T} \tag{5.1.2}$$

Eq. (5.1.2) is identical to Eq. (2.2.4), the basic formula for VMM; hence:

$$P(Q_T) = P(Q_T \mid \text{Model}) = P(q_1, q_2, \ldots q_T \mid \Lambda_1) = P(q_T \mid U(q_{[T:1]}) \mid \Lambda_1) \tag{5.1.3}$$

From Eq. (5.1.2), we can derive the marginal probability for $q_T$:

$$P(q_T) = \sum_{q_1,\cdots q_{T-1}} P(q_1, q_2, \ldots q_T) = \sum_{q_1,\cdots q_{T-1}} \pi_{q_1} \cdot a_{q_1,q_2} \cdots a_{q_{T-1},q_T} \tag{5.1.4}$$

Eq. (5.1.4) is identical to Eq. (1.2.19), our expanded MEF for an HMC:

$$P(q_T) = P(q_T \mid \Lambda_T) = P(q_T \mid \tilde{A}^{T-1} \mid \Lambda_1) \tag{5.1.5}$$

$$\mid \Lambda_T) = \tilde{A}^{T-1} \mid \Lambda_1) = \sum_{q_1,\cdots q_{T-1}} U(q_{[T,1]}) \mid \Lambda_1) \tag{5.1.6}$$

## 5.2. The HMM as a DBN of 2*T*-Nodes

If we take the *T* states in the sequential event space in Eq. (2.2.1) and the *T* observed values in Eq. (3.2.1) as a 2*T*-node DBN, then its graph in PGM is shown in Fig. 5.2.1 below.

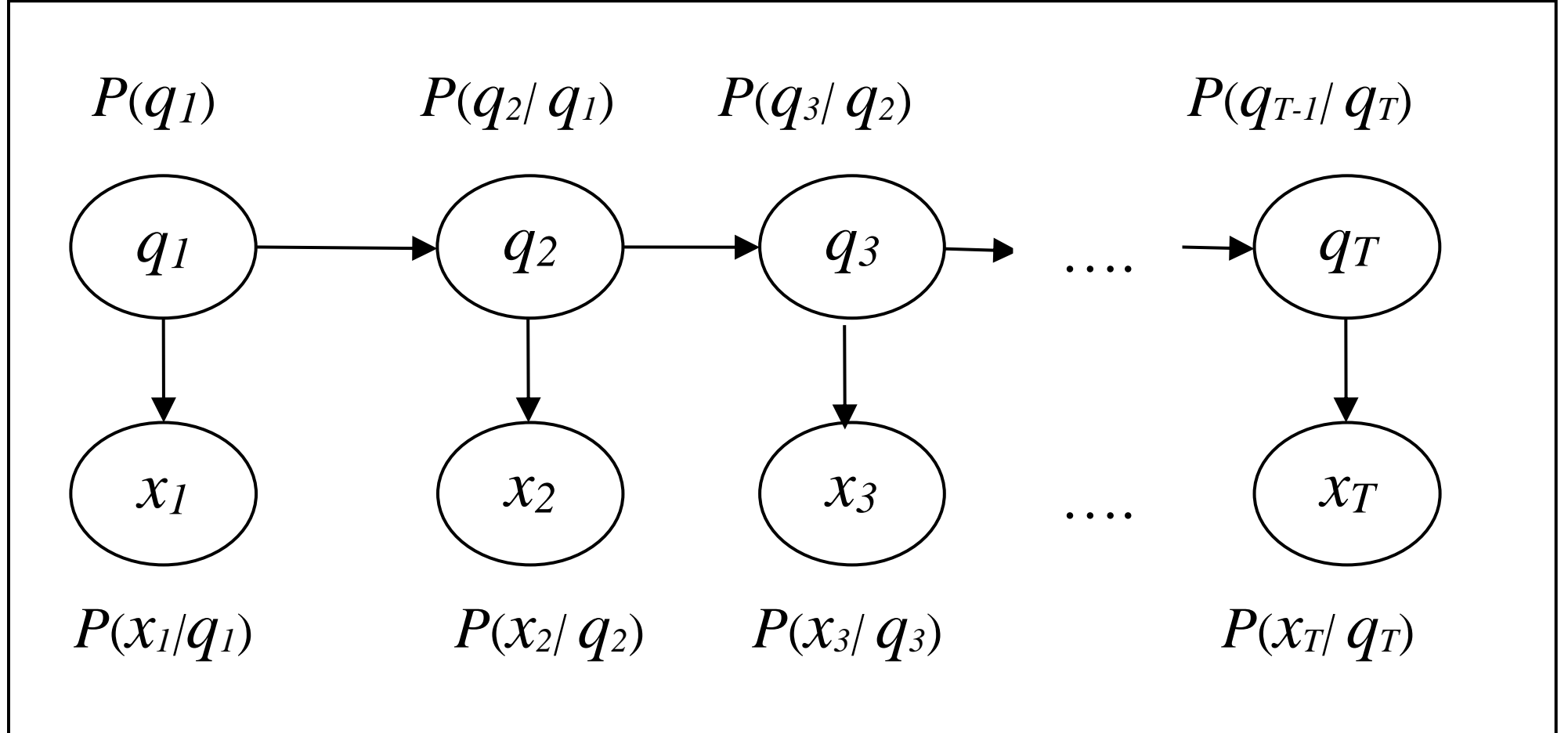

**Fig. 5.2.1**: A Hidden Markov Model as a Dynamic Bayesian Network

Reading from the graph, we have the following full JP:

$$P(x_1,\ldots,x_T; q_1,\ldots q_T) = P(q_1)P(q_2 \mid q_1)P(x_1 \mid q_1)\cdots P(q_T \mid q_{T-1})\,P(x_T \mid q_T) \tag{5.2.1}$$

Applying Eq. (1.2.5), (1.2.10) and (3.2.3a), it can be expressed as:

$$P(X_T, Q_T) \equiv P(x_1, ..., x_T; q_1, ...q_T) = \pi_{q_1} \cdot b_{q_1, x_1} \cdot a_{q_1, q_2} \cdot b_{q_2, x_2} \cdots b_{q_T, x_T} \cdot a_{q_{T-1}, q_T} \quad (5.2.2)$$

Eq. (5.2.2) is identical to Eq. (3.2.8), the FJP of a given observed sequence $X_{[1:T]}$ and a given state sequence $Q_{[1:T]}$.

$$\begin{aligned} &P(X_T, Q_T) = P(X_T, Q_T \mid \Xi_1) \underset{(3.2.8)}{=} P(X_T \mid Q_T) \cdot P(Q_T \mid \Xi_1) \\ &\underset{(5.1.3)}{=} P(X_T \mid Q_T) \cdot P(q_T \mid U(q_{[T:1]}) \mid \Xi_1) \end{aligned} \quad (5.2.3)$$

Recall that we have derived the expression for the probability of observed sequence $X_{[1:T]}$ in Eq. (3.2.12) by inserting the identity operator $I_Q$ into the marginal DF of $X$:

$$P(X \mid \Xi_1) = P(X_T \mid I_Q \mid \Xi_1) \underset{(3.2.11)}{=} \sum_{all\ Q_T} P(X_T \mid Q_T) P(Q_T \mid \Xi_1) = \sum_{all\ Q_T} P(X_T, Q_T \mid \Xi_1) \quad (5.2.4)$$

From Eq. (1.4.8) and (5.2.3), the FJP of a given Markov state sequence $Q_T$ and the time evolution of the model in both VMM and HMM can be presented in MSP as:

$$P(Q_T) = P(Q_T \mid \text{Model}) \equiv P(q_1, q_2, \ldots q_T \mid \Lambda_1) = P(q_T \mid U(q_{[T:1]}) \mid \Lambda_1) \equiv P(q_T \mid \Lambda_{Q_T}) \quad (5.2.5)$$

$$\mid \Lambda_{Q_T}) \equiv U(q_{[T:1]}) \mid \Lambda_1) = U(q_{T-1}) \cdots U(q_1) \mid \Lambda_1) \quad (5.2.6)$$

## 5.3. The Factorial HMM and the DBN

Numerous variants (or extensions) of HMMs can be presented as DBNs. One of the variants is called Factorial HMM (FHMM). A simple example [11] is shown in Fig. (5.3.1) below. In this example, the state is described by a joint random variable (or a random vector [1] [2]) of three independent components.

As seen in the graph, the time evolution chain of each component is independent of each other. Therefore, the joint distribution of the state vector $\boldsymbol{X}$ is the product of the three joint distributions; the P-basis transformation from the state vector to the observed variable is also a product of the transformation of each component. This explains why such a HMM is called a factorial HMM. We now express the FJP of the observed $Y$-sequence in terms of matrices $\boldsymbol{A}$ and $\boldsymbol{B}$ (for simplicity, we removed the subscript $T$ from $\boldsymbol{X}$ and $Y$):

$$P(Y \mid \Xi_1) = P(Y \mid I_{\boldsymbol{X}} \mid \Xi_1) = \sum\nolimits_{all\ \boldsymbol{X}} P(Y \mid \boldsymbol{X}) P(\boldsymbol{X} \mid \Xi_1) \quad (5.3.1)$$

$$\begin{aligned} &P(\boldsymbol{X} \mid \Xi_1) = \prod\nolimits_{i=1}^{3} P(X^i \mid \Xi_1) = \prod\nolimits_{i=1}^{3} P(x_T^i \mid U(x_{T:1}^i) \mid \Xi_1) \\ &= \prod\nolimits_{i=1}^{3} \pi^i{}_{x^i} a^i{}_{x^i_1, x^i_2} \cdots a^i{}_{x^i_{T-1}, x^i_T} \end{aligned} \quad (5.3.2)$$

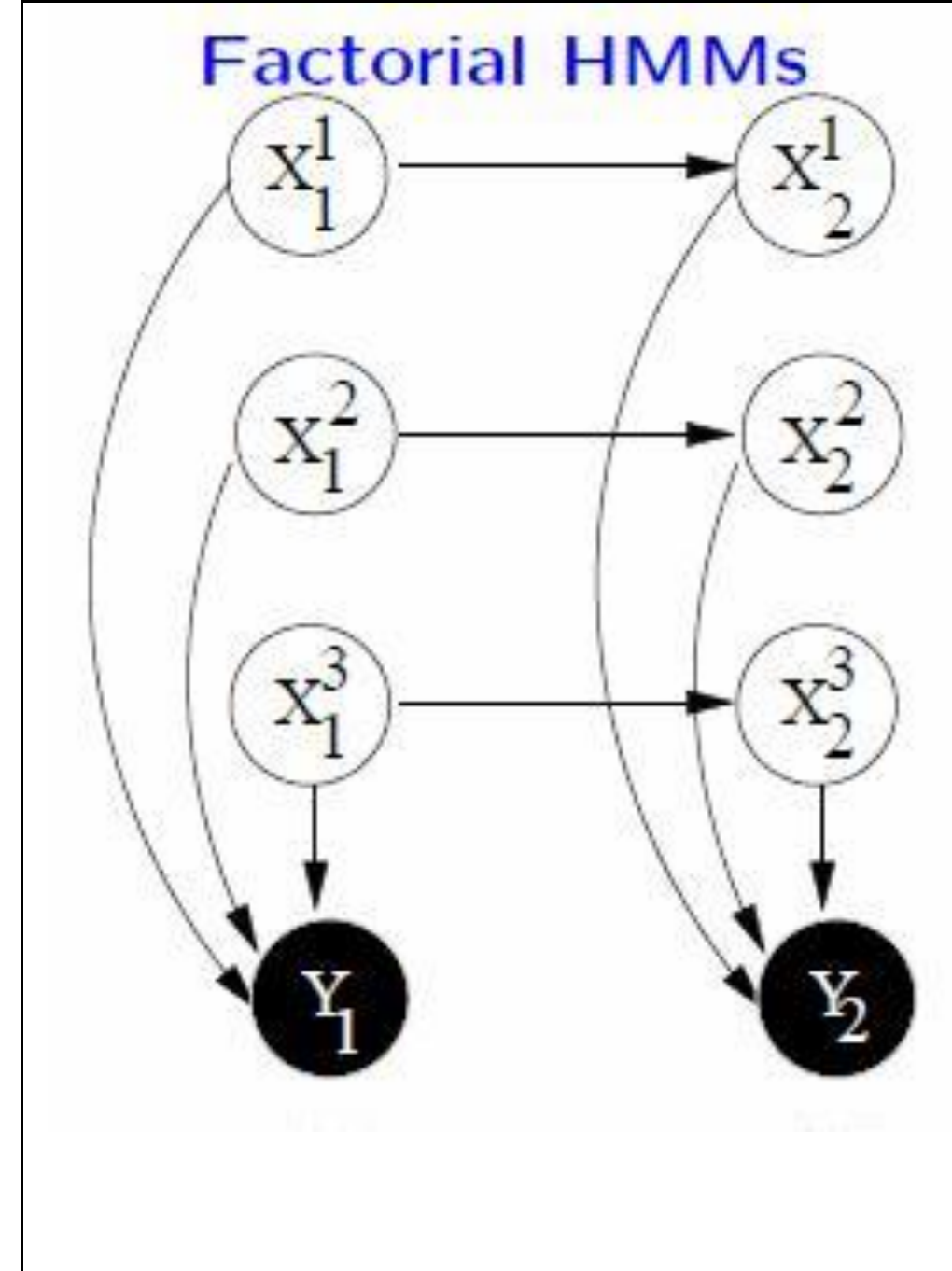

**FHMM Notations**

The observed variable: $Y$

The hidden variable: $\boldsymbol{X} = \{X^1, X^2, X^2\}$

The Markov transition matrix (at time = $t$):

$$P(\boldsymbol{x}_{t+1} \mid \boldsymbol{x}_t) = \prod_{i=1}^{3} P({x_{t+1}}^i \mid {x_t}^i) = \prod_{i=1}^{3} a^i_{{x^i}_t, {x^i}_{t+1}}$$

The FJP of the observed state chain:

$$P(\boldsymbol{X}_2 \mid \lambda_1) = \prod_{i=1}^{3} P(X^i_2 \mid \Xi_1) = \prod_{i=1}^{3} P(x^i_2 \mid U(x^i_1) \mid \Xi_1)$$

$$= \prod_{i=1}^{3} P(x^i_2 \mid x^i_1) P(x^i_1 \mid \Xi_1) = \prod_{i=1}^{3} a^i_{x^i_1 x^i_2} \pi^i_{x^i_1}$$

The basis transformation matrix (at time = t):

$$P(y_t \mid \boldsymbol{x}_t) = \prod_{i=1}^{3} P(y_t \mid {x_t}^i) = \prod_{i=1}^{3} b^i_{{x^i}_t, y_t}$$

The unit operator of $\boldsymbol{X}$:

$$\boldsymbol{I}_X = \sum_X \mid \boldsymbol{X}) P(\boldsymbol{X} \mid$$

**Fig. 5.3.1**: A Factorial HMM [11] as a Dynamic Bayesian Network

$$P(Y \mid \boldsymbol{X}) = \prod_{i=1}^{3} P(Y \mid X^i) = \prod_{i=1}^{3} \prod_{t=1}^{T} b^i_{{x^i}_t, y_t} \tag{5.3.3}$$

The joint probability can then be expressed as the product of Eq. (5.3.3) and (5.3.2):

$$P(Y, \boldsymbol{X}) = P(Y, \boldsymbol{X} \mid \Xi_1) \underset{(3.2.7)}{=} P(Y \mid \boldsymbol{X}) \cdot P(\boldsymbol{X} \mid \Xi_1) \tag{5.3.4}$$

According to Eq. (5.3.2), we can represent the evolution of the hidden Variables $\boldsymbol{X}$ as:

$$\mid \Xi_{X_T}) = \prod_{i=1}^{3} U(x^i_{T:1}) \mid \Xi_1) = \prod_{i=1}^{3} \mid x^i_T)\, {\pi^i}_{x^i} {a^i}_{{x^i}_1, {x^i}_2} \cdots {a^i}_{{x^i}_{T-2}, {x^i}_{T-1}} \tag{5.3.5}$$

By observing Eqs. (1.4.8), (2.3.12), (5.2.6), and (5.3.5), we see that the time evolution of a Markov chain or a sequence in HMC, VMM, HMM, and FHMM has been unified through the Markov P-projector sequence (MSP):

$$\mid \Omega_T) = \sum_{q_1, \cdots q_{T-1}} U(q_{[T:1]}) \mid \Omega_1), \quad \mid \Lambda_{Q_T}) = U(q_{[T:1]}) \mid \Lambda_1), \quad \mid \Xi_{X_T}) = U(\boldsymbol{X}_{[T:1]}) \mid \Xi_1) \tag{5.3.6}$$

In Section 6, we will show that Eq. (5.3.6) can also be applied to continuous-time Markov processes (CT-MP).

## 5.4. The Feedback in Extended HMMs of DBNs

A serious issue of static Bayesian networks is that they do not allow for feedback [16]. However, the feedback mechanism can be implemented in extended HMMs [17, Fig. 3], as a subclass of DBN.

Fig. (5.4.1) is the graph of a simple extended HMM, which allows feedback from a child node at time *t* to its parent node at time *t* + 1. Since the transition probability from $q_1$ to $q_2$ of the Markov sequence is replaced by the conditional probability $P(q_2 \mid q_1, x_1)$, the extended HMM is not a standard hidden Markov model anymore.

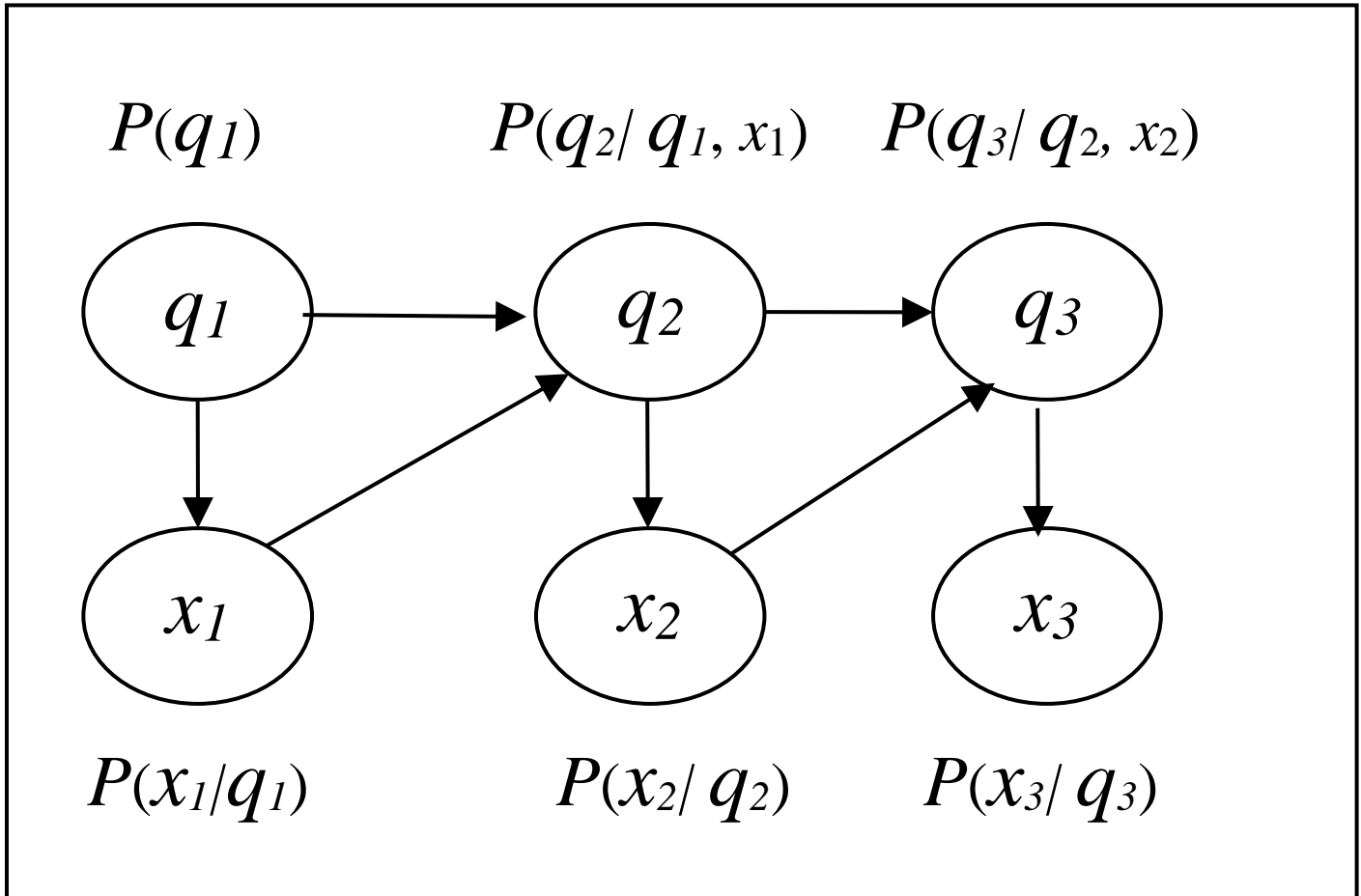

**Fig. 5.4.1**: Feedback in an Extended HMM

From Fig. 5.4.1, we can readily write down the full JP:

$$\begin{aligned} &P(X_T;Q_T) = P(q_1)P(x_1 \mid q_1)P(q_2 \mid q_1, x_1)P(x_2 \mid q_2)P(q_3 \mid q_2, x_2)\,P(x_3 \mid q_3) \\ &= \pi_{q_1} \cdot b_{q_1,\,x_1} \cdot c_{q_1,x_1;q_2} \cdot b_{q_2,\,x_2} \cdot c_{q_2,x_2;q_3} \cdot b_{q_3,\,x_3}; \quad c_{q_{t+1},x_t;q_t} \equiv P(q_{t+1} \mid q_t, x_t) \end{aligned} \tag{5.4.1}$$

The Viterbi algorithm in Sec. 3 can be adapted for this extended HMM. Only the recursion Eq. (3.3.7) needs a minor change (replacing elements $a_{i,j}$ with $c_{i,x;j}$):

$$\begin{aligned} &\delta_t(j) = \max_{1 \le i \le N} \left\{ \delta_{t-1}(i) \cdot c_{i,x;\,j} \right\} \cdot b_{j,\,x_t}, \quad 2 \le t \le T,\ 1 \le j \le N, \quad x \equiv x_{t-1} \\ &\psi_t(j) = \arg\max_{1 \le i \le N} \left\{ \delta_{t-1}(i) \cdot c_{i,x;\,j} \right\}, \quad 2 \le t \le T,\ 1 \le j \le N, \quad x \equiv x_{t-1} \end{aligned} \tag{5.4.2}$$

For the Weather-Stone example, the 3×3 table (1.2.1) is now replaced by a 6×3 table (today's weather and stone state vs tomorrow's weather).

## 6. PBN and Continuous-Time Markov Processes

The unified evolution equations in Eq. (5.3.6) can also be applied to continuous-time Markov processes (**CT-MP**) with infinite discrete states. The stochastic process can be represented by:

$$X(t) \to x_i, t \in [0,\infty) \subset \Re,\ i, x_i \in \mathbb{N},\ p_{i,j}(t,s) = p(i,t\,|\,j,s) \equiv P(X(t)=i\,|\,X(s)=j) \tag{6.1}$$

The transition probability[4] of a homogeneous CT-MP has the following property [1, Eq. (76)], [18, Sec. 3.3]:

$$p_{j,i}(t+s,s) \equiv P(x_i,t+s\,|\,x_j,s) \underset{\text{HMC}}{=} P(x_i,t\,|\,x_j,0) \equiv p_{j,i}(t), \quad p_j(t) \equiv p_{j,0}(t) \tag{6.2}$$

Using the Kolmogorov Backward equations (see PBN's derivation in [1, Sec 3.5]), a CT-MP has the following differential-difference equations (DDEs) [18]:

$$p'_{i,j}(t) = \sum_k g_{ik}\, p_{k,j}(t), \quad \underset{(6.2)}{\leftrightarrow} \quad p'_i\,(t) = \sum_k g_{ik}\, p_k(t), \quad \underset{\text{PBN}}{\leftrightarrow} \quad \partial_t\,|\,\Omega_t) = G\,|\,\Omega_t) \tag{6.3}$$

Here, $G$ is called the generator matrix, and its elements are given by [19]:

$$\begin{aligned} & g_{i\,j} = \lim_{h\to 0^+}\left[(p_{i\,j}(h) - \delta_{i\,j})/h\right] = h'_{i\,j}(0) \\ & \textstyle\sum_j p_{i,j} \underset{(1.2.4)}{=} 1 \to \sum_j p'_{i,j} = 0 \to g_{i,i} = \sum_{j\neq i} g_{i,j} \end{aligned} \tag{6.4}$$

**CT-VMM**: Assuming we have solutions for the DDEs in Eq. (6.3), even though only a few CT-MPs, such as the Poisson process [1], [18], possess exact solutions, we can express the full joint probability for the VMM in Fig. 2.1.1 or Eq. (2.1.9) with T=3 as:

$$\begin{aligned} & P(q_3\,|\,Q_T) = P(q_3\,|\,U(q_{[3:1]})\,|\,\Omega_1) = P(q_3\,|\,q_2)P(q_2\,|\,q_1)P(q_1\,|\,\Omega_1) \\ & = p_{q_2,q_3}(t_3 - t_2)\,p_{q_1,q_2}(t_2 - t_1)\cdot \pi_{q_1}; \quad P(q_i\,|\,q_j) \equiv P(q_i,t_i\,|\,q_j,t_j) \end{aligned} \tag{6.5}$$

If the HMC is a *Poisson process*, we can use Eq. (85) from reference [1]:

$$p_{i,j}(t) = P(j,t\,|\,i,0) = \frac{(\lambda t)^{j-i}}{(j-i)!}e^{-\lambda t}, \text{if } \ j \geq i; \quad p_{i,j}(t) = 0, \text{ if } j < i \tag{6.6}$$

Summing over all intermediate states, we get the MEF of the HMC, similar to Eq. (1.4.7):

$$P(q_3\,|\,\Omega_t) = \textstyle\sum_{q_2,q_1} p_{q_2,q_3}(t_3)\,p_{q_1,q_2}(t_2)\cdot \pi_{q_1} \equiv P(q_3\,|\,\tilde{P}(t_3)\tilde{P}(t_2)\,|\,\Omega_1) \tag{6.7}$$

**CT- HMM:** Suppose that the hidden Markov sequence $Q_T$ and the observation sequence $X_T$ are given in Eq. (5.2.1) or Fig. (5.2.1) with $T$=3, we have the following full JP:

[4] The transition matrix element may be denoted as the transport of the matrix in our definition.

$$
\begin{aligned}
P(X_T, Q_T) &= P(q_1)P(q_2 \mid q_1)P(x_1 \mid q_1)\,P(q_3 \mid q_2)\,P(x_3 \mid q_3) \\
&= \pi_{q_1} \cdot b_{q_1,\,x_1} \cdot p_{q_1,q_2}(t_2) \cdot b_{q_2,\,x_2} \cdot p_{q_2,q_3}(t_3) \cdot b_{q_3,\,x_3}
\end{aligned} \tag{6.8}
$$

**CT-HMM with Continuous Random Variables**: Eq. (6.3) can be extended to a continuous random variable. The generator matrix becomes a *generator operator* now, and the equation becomes a *master equation* [1, Eq. (62)]:

$$
p'(x,t) = \int dx' \hat{G}(x,x')\,p(x',t) \quad \underset{\text{PBN}}{\leftrightarrow} \quad \partial_t \,|\,\Omega_t) = \hat{G}\,|\,\Omega_t) \tag{6.9}
$$

For a HMC ( $\partial_t G = 0$ ), it has the following symbolic solution [1, Eq. (62) and (87)]:

$$
|\,\Omega_t) = e^{\hat{G}(t-t')}\,|\,\Omega_{t'}), \quad P(x,t \mid x',t') = P(x \,|\, e^{\hat{G}(t-t')} \,|\, x'), \quad t \geq t' \tag{6.10}
$$

The FJP of the Markov state sequence (MSS) in Eq. (6.5) of a VMM, or Eq. (6.8) of an HMM, reads:

$$
\begin{aligned}
P(q_3 \mid Q_T) &= P(q_3 \,|\, U(q_{[3:1]}) \,|\, \Omega_1) = P(q_3 \mid q_2)P(q_2 \mid q_1)P(q_1 \mid \Omega_1) \\
&= P(q_3 \,|\, e^{\hat{G}(t_3-t_2)} \,|\, q_2)P(q_2 \,|\, e^{\hat{G}(t_2-t_1)} \,|\, q_1)P(q_1 \,|\, \Omega_1)
\end{aligned} \tag{6.11}
$$

If the HMC is an *Einstein-Brown process*, we can apply Eq. (108) of [1]: (6.12)

$$
P(q_b \mid q_a) \equiv P(x_b, t_b \mid x_a, t_a) = \sqrt{\frac{1}{4\pi D(t_b - t_a)}} \exp\left[ -\frac{(x_b - x_a)^2}{4D(t_b - t_a)} \right] \tag{6.13}
$$

Remember that the observation of an HMM corresponds to inserting an appropriate P-projector into the Markov evolution process, regardless of whether the evolution time is discrete or continuous. By integrating over the intermediate states in Eq. (6.11), we return to the MEF of an HMC, as shown in Eq. (6.10):

$$
\begin{aligned}
P(q_3 \,|\, \Omega_t) &= \iint dq_2 dq_1 P(q_3 \,|\, e^{\hat{G}(t_3-t_2)} \,|\, q_2)P(q_2 \,|\, e^{\hat{G}(t_2-t_1)} \,|\, q_1)P(q_1 \,|\, \Omega_1) \\
&= P(q_3 \,|\, e^{\hat{G}(t_3-t_2)} I_Q e^{\hat{G}(t_2-t_1)} I_Q \,|\, \Omega_1) = P(q_3 \,|\, e^{\hat{G}(t_3-t_1)} \,|\, \Omega_1), \quad I_Q = \int |\, q) dq (q \,|
\end{aligned} \tag{6.14}
$$

Certainly, more work remains to be done. We only present the PBN’s preliminary formulation of CT-VMMs and CT-HMMs here. Readers can refer to lecture notes and papers discussing CT-HMCs, CT-HMMs, and related algorithms [18-22].

## 7. Summary and Discussion

The P-projectors and Markov sequence projectors (MSP) were introduced, highlighting the close relationship between the full joint probability (FJP) of a visible Markov model (VMM) and the expanded Markov evolution formula (MEF). The P-bases of the hidden Markov sequence and the observations in the sequential event space are defined for hidden Markov models (HMM), whose basic formulas were interpreted by the MSP and the transformations between the two P-bases. The Viterbi algorithm was methodically applied to the well-known weather-stone HMM example. The Java software package Elvira[5] was utilized to illustrate the graphical structure and confirm the results of the Viterbi algorithm applied to the weather-stone HMM example. The factorial HMM, the extended HMM with feedback, and the continuous-time VMMs and HMMs with discrete or continuous states are briefly discussed. All the above Markov models fall under dynamic Bayesian networks (DBNs).

In Ref [12], we demonstrated that P-identities provide a simplified method for managing various probabilities in static Bayesian networks. This article illustrates how P-identities, P-projectors, and MSPs can be applied to different Markov models, thereby simplifying and unifying the time evolution expressions of system P-kets for Markov state sequences (MSSs). Our study validated the potential of PBN as a valuable alternative tool for managing multivariable systems in both static and dynamic Bayesian networks, which are essential in Machine Learning (ML) and Artificial Intelligence (AI) [23].

**Abbreviations:**

| | |
|---|---|
| BN: | Bayesian Network, Dynamic BN (DBN), Static BN (SBN) |
| CT-HMM | Continuous-Time Hidden Markov Model |
| CT-MP: | Continuous-Time Markov Process |
| FHMM: | Factorial Hidden Markov Model |
| FJP: | Full Joint Probability |
| HMM: | Hidden Markov Model |
| HMC: | Homogenous Markov Chain |
| JP: | Joint Probability |
| MEF: | Markov Evolution Formula |
| MSP: | Markov Sequence Projector |
| MSS: | Markov State Sequence |
| P-basis: | Probability-basis, P-bra, P-ket, P-identity, P-projector |
| PBN: | Probability Bracket Notation |
| PGM: | Probabilistic Graphic Model |
| SES: | Sequential Event Space |
| V-basis: | Vector-basis, V-bra, V-ket |
| VMM: | Visible Markov Model |

[5] The Elvira software is not available online anymore. One may try tools such as the Bayes Server [23].